\documentclass[preprint,12pt]{elsarticle}

\usepackage{amssymb}
\usepackage{amsmath}
\usepackage{graphics} 
\usepackage{epsfig} 
\usepackage{mathptmx} 
\usepackage{times} 
\usepackage{amsmath} 
\usepackage{amssymb}  
\usepackage{graphicx}
\usepackage{bm}
\usepackage{eucal}
\usepackage{todonotes}[inline]  

\usepackage{float}
\usepackage[ruled, lined, linesnumbered, commentsnumbered, longend]{algorithm2e}
\usepackage{pgfplotstable}
\usepackage{pgfplots}
\usepackage{tikz}
\usetikzlibrary[pgfplots.groupplots]
\usepgfplotslibrary{fillbetween}
\pgfplotsset{compat=1.5}
\usepackage{subcaption}
\usepackage{caption}
\usepackage{xcolor}
\usepackage{enumerate}
\usepackage{graphicx,import}
\usepackage{makecell}
\usepackage{wrapfig}
\usepackage{colortbl}

\usepackage{graphicx}
\usepackage{tabularx,ragged2e,booktabs}
\usepackage{url}
\usepackage{siunitx}
\usepackage{rotating}
\usepackage{amssymb}
\usepackage{tabularx}
\usepackage{pgfplots}
\usepackage{csvsimple}
\usepackage{multirow}
\usepackage{float}
\usepackage{graphicx}
\usepackage{booktabs}
\usepackage{svg}
\usepackage{cite}
\usepackage{gensymb}

\SetKwComment{Comment}{$\triangleright$\ }{}

\SetCommentSty{mycommfont}

\newcommand\BibTeX{{\rmfamily B\kern-.05em \textsc{i\kern-.025em b}\kern-.08em
T\kern-.1667em\lower.7ex\hbox{E}\kern-.125emX}}

\journal{Robotics and Autonomous Systems}

\begin{document}

\begin{frontmatter}




\title{ArtReg: Visuo-Tactile based Pose Tracking and Manipulation of Unseen Articulated Objects}


\author[bmw,uog]{Prajval Kumar Murali}
\author[bmw,tue]{Mohsen Kaboli}

\affiliation[bmw]{organization={RoboTac Lab, BMW Group}}
\affiliation[uog]{organization={University of Glasgow}}
\affiliation[tue]{organization={Eindhoven University of Technology}}

\begin{abstract}
Robots operating in real-world environments frequently encounter unknown objects with complex structures and articulated components, such as doors, drawers, cabinets, and tools. The ability to perceive, track, and manipulate these objects without prior knowledge of their geometry or kinematic properties remains a fundamental challenge in robotics. 
In this work, we present a novel method for visuo-tactile-based tracking of unseen objects (single,
multiple, or articulated) during robotic interaction without assuming any prior knowledge regarding object shape or dynamics. Our novel pose tracking approach termed \textit{ArtReg} (stands for Articulated Registration) integrates visuo-tactile point clouds in an unscented Kalman Filter formulation in the SE(3) Lie Group for point cloud registration. ArtReg is used to detect possible articulated joints in objects using purposeful manipulation maneuvers such as pushing or hold-pulling with a two-robot team. Furthermore, we leverage ArtReg to develop a closed-loop controller for goal-driven manipulation of articulated objects to move the object into the desired pose configuration. We have extensively evaluated our approach on various types of unknown objects through real robot experiments. We also demonstrate the robustness of our method by evaluating objects with varying center of mass, low-light conditions, and with challenging visual backgrounds. Furthermore, we benchmarked our approach on a standard dataset of articulated objects and demonstrated improved performance in terms of pose accuracy compared to state-of-the-art methods.
Our experiments indicate that robust and accurate pose tracking leveraging visuo-tactile information enables robots to perceive and interact with unseen complex articulated objects (with revolute or prismatic joints).
\end{abstract}




\end{frontmatter}

\section{Introduction}
For robots to work in dynamic environments, both perceiving and manipulating complex objects are often necessary. Articulated objects are ubiquitous in our surroundings, exemplified by items such as cabinet drawers, eyeglass frames, scissors, laptops, etc. Pose tracking of articulated objects is challenging due to the high dimensionality of the state space, stemming from the multiple degrees of freedom and the inherent nonlinearity in dynamics, wherein the motion imparted to one component nonlinearly influences the movement of interconnected parts. Furthermore, interactive perception, where purposeful manipulation actions are performed to improve perception, is often necessary to detect possible articulation joints and correctly manipulate objects without damaging them. Hence, there is a clear need for augmenting visual perception with tactile sensing capable of discerning contact force and location. Pose tracking algorithms that can leverage multimodal sensing data and operate without prior knowledge of object models or kinematic properties can enhance the versatility and adaptability of robotic systems. 

\begin{figure}[]
    \centering
    \includegraphics[width = \columnwidth]{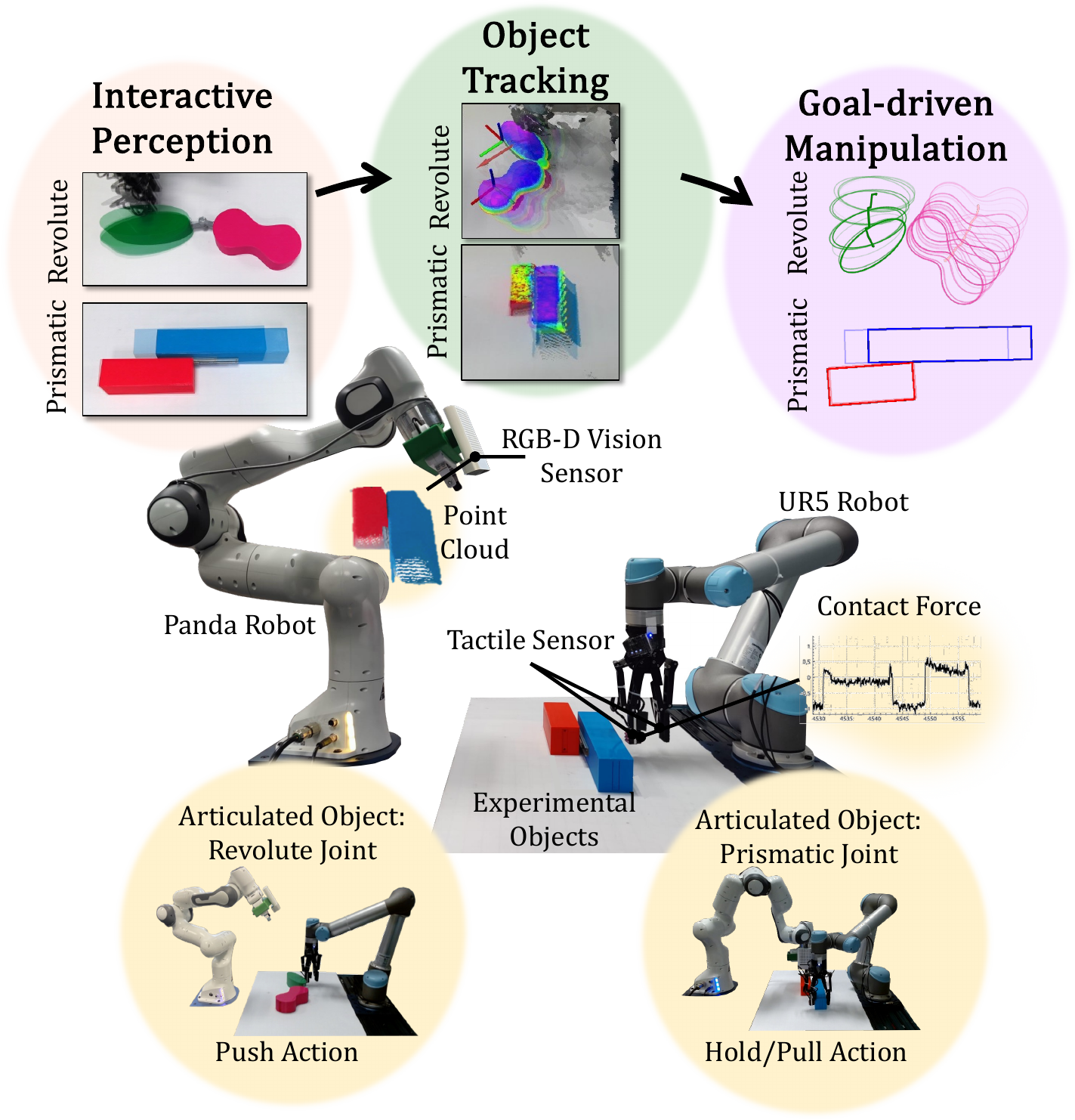}
    \caption{Experimental setup: A Franka Emika Panda robot with a Azure Kinect DK RGB-D vision sensor and a Universal Robots UR5 sensorised with tactile sensor arrays on the Robotiq Gripper, with unknown articulated objects in the workspace. The robots perform interactive perception to detect possible articulation structure in the objects. The objects are tracked using our ArtReg algorithm and the 6 degree-of-freedom (DoF) tracking information is used for goal-driven manipulation.}
    \label{fig:fig1}
\end{figure}

Several prior works on the tracking of articulated objects have fundamentally relied on geometric feature tracking or the utilization of marker-based methodologies applied to sequential image data~\citep{martin2022coupled,sturm2011probabilistic}. Other works have also assumed prior knowledge of the object models or the kinematic structure of the articulated objects for tracking and manipulation~\citep{nickels2001model, paolillo2018interlinked}. More recent works have used factor-graph based approaches to estimate articulated object structure from input RGB-D streams~\citep{heppert2022category}.
However, the predominant focus has been on visual inputs, overlooking the potential benefits of incorporating tactile information.
Tactile information can provide complementary information regarding the object properties and is invariant to occlusions, ambient lighting, and object transparency~\citep{murali2024shared, Qiang-TRO-2020}. Furthermore, contact force information is useful during interactive perception to detect possible articulation and joint constraint limits without damaging the objects.
Real-time tracking of unseen objects without prior knowledge regarding articulation constraints is an open research problem. Pose tracking not only enables precise object localization but also facilitates downstream tasks like goal-driven manipulation. Most prior works performing closed-loop control for goal-driven manipulation uses only rigid objects or disregards proprioceptive or tactile information for the task~\citep{lloyd2021goal, paolillo2018interlinked,katz2008manipulating, xu2022universal}. 
Adapting such techniques directly to articulated objects presents a non-trivial undertaking, necessitating novel approaches tailored to the complexities inherent in articulated structures.

Tackling these current challenges in the state-of-the-art, we propose a novel framework for visuo-tactile-based interactive perception for tracking and manipulating unknown novel objects (single, multiple, and articulated with revolute or prismatic joints) without assuming any prior knowledge regarding object shape or dynamics. 
Our contributions are as follows:
\begin{enumerate}[I]
    \item We propose ArtReg, a novel method for tracking unknown novel objects (single, multiple, or articulated) by integrating visual and tactile point cloud data with a Manifold Unscented Kalman Filter on the SE(3) Lie Group.
    \item Based on ArtReg, we develop a novel approach for the detection of kinematic chains (revolute or prismatic joints) in unknown objects using a combination of manipulation actions facilitated by autonomous interactive visuo-tactile perception.
    \item Further leveraging ArtReg, we develop a visuo-tactile based closed-loop control algorithm intended for precise manipulation of objects to a goal configuration. 
    \item We perform extensive experiments to validate the proposed framework under a variety of conditions, including low-light environments, complex backgrounds, and variations in the center-of-mass of the objects.
\end{enumerate}
We perform extensive experiments with a two robot setup shown in Fig.~\ref{fig:fig1} where one robot is equipped with a RGB-D visual sensor and another robot is sensorised with tactile sensor arrays on the gripper.
This paper is organized as follows: Sec.~\ref{sec:relatedwork} summarizes the state-of-the-art in interactive perception, articulated object tracking and goal-driven manipulation and highlights our contributions in
the context of current and related research. Our
methodology is presented in Sec.~\ref{sec:methods}.
Experimental results are reported in Sec.~\ref{sec:experiments} and finally concluded in Sec.~\ref{sec:conclusion}.

\section{Related Work}
\label{sec:relatedwork}
Our work focuses on the detection, tracking, and closed-loop control of various types of objects, in particular articulated objects. Hence, in this section we review the state-of-the-art works in each domain and their relation to our work.

There are a number of computer vision approaches for tracking articulated objects using 2D/ 3D information from vision-based sensors~\citep{lowe1991fitting, nickels2001model, pellegrini2008generalisation, schmidt2015dart}.
Early works used model-based methods with known object models and kinematic structure to track articulated objects using least-squares optimization techniques~\citep{lowe1991fitting} or Kalman filters~\citep{nickels2001model}. More recently, Schmidt et al.~\citep{schmidt2015dart} presented DART, which uses the signed distance function to track articulated objects using off-the-shelf depth sensors wherein the model of the objects needs to be known \textit{a priori}. Similarly, extensions of classical point cloud registration techniques, such as iterative closest point (ICP) for articulated objects have also been explored~\citep{pellegrini2008generalisation}.
Similarly, Michel et al.~\citep{michel2015pose} proposed a RANSAC-based approach to generate $K$ correspondences for the $K$-part kinematic chain of articulated objects.
Depth data was combined with joint encoder information with a Kalman filter to track articulated robot bodies~\citep{cifuentes2016probabilistic}. 
Finally, deep learning techniques have also been used for tracking articulated objects and humans pose tracking with colour and depth images~\citep{liu2022toward,kanazawa2018end, ge20193d}.
More recently, Articulation-aware Normalized Coordinate Space Hierarchy (ANCSH) method was introduced to perform category-level pose estimation of articulated objects~\citep{li2020category}. 
Similarly, the authors in~\citep{liu2022toward} proposed part-pair based pose estimation and the joint properties for the connected parts for real-world objects based on input RGB-D images. 
The progression of large Vision-Language Models (VLMs), has facilitated the generation of a diverse range of articulated object models and their corresponding affordances, which are instrumental in training networks for object tracking and manipulation~\citep{le2024articulate}.

Compared to passive computer vision techniques, interactive perception has been employed for detection and tracking of articulated objects using embodied robotic agents.
Katz et al.~\citep{katz2014interactive} developed a framework to recognize articulated objects using predefined robot motions to interact with objects and relying on segmentation and relative object motion analysis to decompose articulated objects into kinematic chains.
Sturm et al.~\citep{sturm2011probabilistic} proposed a probabilistic approach for the identification of the kinematic model of articulated objects. They assume knowledge of the number of links and complete perceptual information, i.e. the pose evolution of each link over time in order to learn about the joints (revolute, prismatic, or rigid) and build the kinematic model.
Considering multimodal perception for tracking articulated objects, Mart\'{i}n-Mart\'{i}n et al.~\citep{martin2022coupled} proposed a hierarchical recursive Bayesian filtering method combining RGB, depth, and force-torque sensing. These works use markers or handcrafted feature extraction methods to track the moving parts in consecutive image frames. Desingh et al.~\citep{desingh2019efficient} designed a nonparametric belief propagation technique for articulated object pose estimation. However, they relied upon known 3D models of articulated parts of the object and the articulation kinematic constraints which limits the applicability of the algorithm to unknown scenarios. Factor graphs have also been used for pose tracking of articulated objects~\citep{heppert2022category}.
Recently, deep learning techniques along with interactive perception have been used for articulation detection. Weng et al.~\citep{weng2021captra} proposed CAPTRA for category-level pose tracking of articulated objects from point clouds. They assume to have link-wise segmented point clouds as well as pose initialization and build canonical models to regress over scale, rotation, and translation for each link of the unknown category-level object. Other works focused on the problem of predicting the articulated motion as well as the articulated axis through observation of point cloud sequences which are used for robot manipulation action planning~\citep{zhang2023flowbot}. The deep learning works utilize category-level information for the estimation of articulation parameters, resulting in a dependency on categories and relevant use-case dependent datasets and training overhead. Gadre et al.~\citep{gadre2021act} presented a policy learning method to extract articulation information from visual input and associated action strategies to isolate and discover parts of the image. Similarly, Mo et al.~\citep{mo2021where2act} described a learning-from-interaction framework in which the network predicts possible action affordances at each pixel when provided with RGB-D images. Furthermore, sequence of RGB images was used to train a network to extract 3D planar surfaces that have articulation joints~\citep{qian2022understanding}.

While interactive perception techniques have primarily been used for detection of kinematic structure in articulated objects, it can also be used with closed-loop control for manipulating the object into a desired goal state. Commonly used manipulation actions involve pick-and-place (prehensile) and pushing (non-prehensile)~\citep{stuber2020let}. For instance, Lloyd and Lepora~\citep{lloyd2021goal} presented a tactile-only goal-driven pushing control framework for single rigid objects without relying on visual feedback. 
Similarly, Dengler et al.~\citep{dengler2022learning} presented a reinforcement learning strategy for goal-driven pushing on a single object in clutter using RGB-D information.
Although there are many works in the literature that study goal-driven pushing or other manipulation actions, they are limited to single rigid objects only in controlled or unstructured environments~\citep{stuber2020let}.
Few works tackled the planning problem for articulated object manipulation. Pflueger et al.~\citep{pflueger2015multi} implemented a hierarchical multi-step planner for manipulating a chair with an articulated revolute joint with a two-arm robot. However, their implementation is very specific as they assume the knowledge of the full model of the chair including the kinematics of the articulated joint, markers for tracking the pose and predefined grasp locations. Mittal et al.~\citep{mittal2022articulated} proposed a two-stage planning architecture to provide object-centric actions to manipulate an articulated object and agent-centric planner for whole-body control of a robotic manipulator on a mobile platform in unstructured environments. They used category-level articulated object representation information from~\citep{li2020category} to estimate the object pose and generate action affordances. However, they only used grasp actions on limited sets of articulated objects involving handles on drawers or cabinet frames. 
Such objects are also used with reinforcement learning (RL) based methods for performing opening/closing actions~\citep{xie2023part}.
Similarly, Schiavi et al.~\citep{schiavi2023learning} developed a sampling-based controller for the task for opening and closing an oven door which comprises of a revolute joint. Their framework requires the precise object pose which is provided by an IMU sensor and also the object model to be known \textit{a priori}.
Unlike prior works, in this article, we present a full-fledged framework to detect, track, and manipulate complex objects (single, or articulated) without assuming knowledge of object shape and dynamics.

\section{Methodology}
\label{sec:methods}
\subsection{Notations}
Throughout the article, we employ the following notations:
\begin{enumerate}
    \item Matrices are expressed in upper-case bold face symbols whereas vectors are expressed in lower-case bold face symbols.
    \item $\mathbb{I}_n \in \mathbb{R}^{n\times n}$ denotes a square identity matrix of dimension~$n$.
    \item $\mathbf{0}_{m\times n} \in \mathbb{R}^{m\times n}$ denotes a zero matrix of dimension~$m\times n$.
    \item $(\cdot)^T$ denotes transpose operation and $(\cdot)^{-1}$ denotes inverse operation.
    \item $\hat{(\cdot)}$ denotes estimated value and $\bar{(\cdot)}$ denotes average value.
    \item ${}^AH_{B} \in SE(3)$ denotes a $4\times 4$ homogeneous transformation matrix which transforms a vector in $B$ coordinate frame into a vector expressed in $A$ coordinate frame.
    \item ${}^AR_{B} \in SO(3)$ denotes a $3\times 3$ rotation matrix which rotates a vector in $B$ coordinate frame into a vector expressed in $A$ coordinate frame.
    \item The Lie Group manifold is denoted by $\mathcal{M}$ which is a differentiable smooth manifold. The tangent space at $\mathcal{X} \in \mathcal{M}$ is denoted by $T_{\mathcal{X}}\mathcal{M}$.
    \item The function $\varphi(\cdot)$ represents the retraction function to transform a variable from tangent space to manifold and $\varphi^{-1}(\cdot)$ denotes the inverse retraction operation to move from the manifold to tangent space. In the case of Lie Groups, $\varphi(\cdot)$ is termed the exponential map (\texttt{exp}) and $\varphi^{-1}(\cdot)$ is the logarithmic map (\texttt{Log}).
    
\end{enumerate}

\begin{figure}
   \centering
   \includegraphics[width = \textwidth]{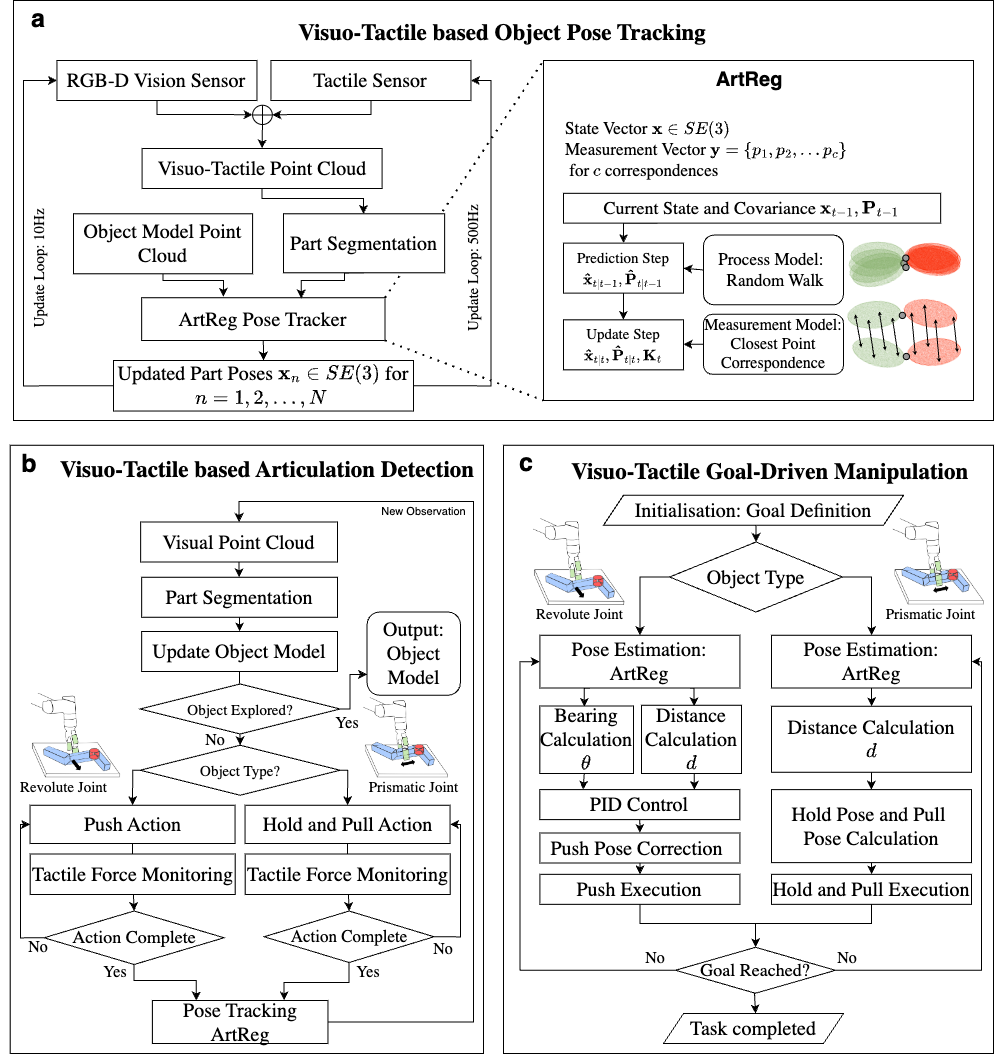}
   \caption{Our proposed framework: (a) Visuo-tactile-based pose tracking method termed ArtReg, (b) Interactive visuo-tactile perception for articulation detection, and (c) Visuo-tactile-based closed-loop control for goal-driven manipulation.}
   \label{fig:framework}
\end{figure}

\subsection{Problem Formulation}
We focus on the problem of pose tracking, object-type detection, and goal-driven manipulation of unseen objects (single, multiple, or articulated) without assuming any prior knowledge regarding object shape or dynamics shown in Fig.~\ref{fig:framework}. 
We present a novel method termed ArtReg for the accurate pose tracking of objects using manifold unscented Kalman filter as shown in Fig.~\ref{fig:framework}a and Sec.~\ref{sec:artreg_mukf}. 
The two-robot team shown in Fig.~\ref{fig:fig1} autonomously perform manipulation actions to detect possible articulation and infer the underlying kinematics if multiple objects are present in the workspace using action primitives as described in Fig.~\ref{fig:framework}b and Sec.~\ref{sec:articulation_deetction}. The final part of the framework (Fig.~\ref{fig:framework}c) presents a closed-loop goal-driven manipulation approach for manipulating articulated objects (with revolute or prismatic joints) or single objects to a desired goal-pose by relying upon visual and tactile sensing described in Sec.~\ref{sec:goal-driven-pushing}.

\subsection{SE(3) Lie Group-based Registration (ArtReg) for Visuo-Tactile Pose Tracking}
\label{sec:artreg_mukf}
We cast the pose tracking problem with point cloud registration as a recursive non-linear Bayesian estimation problem. In particular, we develop a Manifold Unscented Kalman Filter (UKF) formulation on the SE(3) Lie Group for pose tracking. 

\begin{figure}[t!]
    \centering
    \includegraphics[width = 0.9\textwidth]{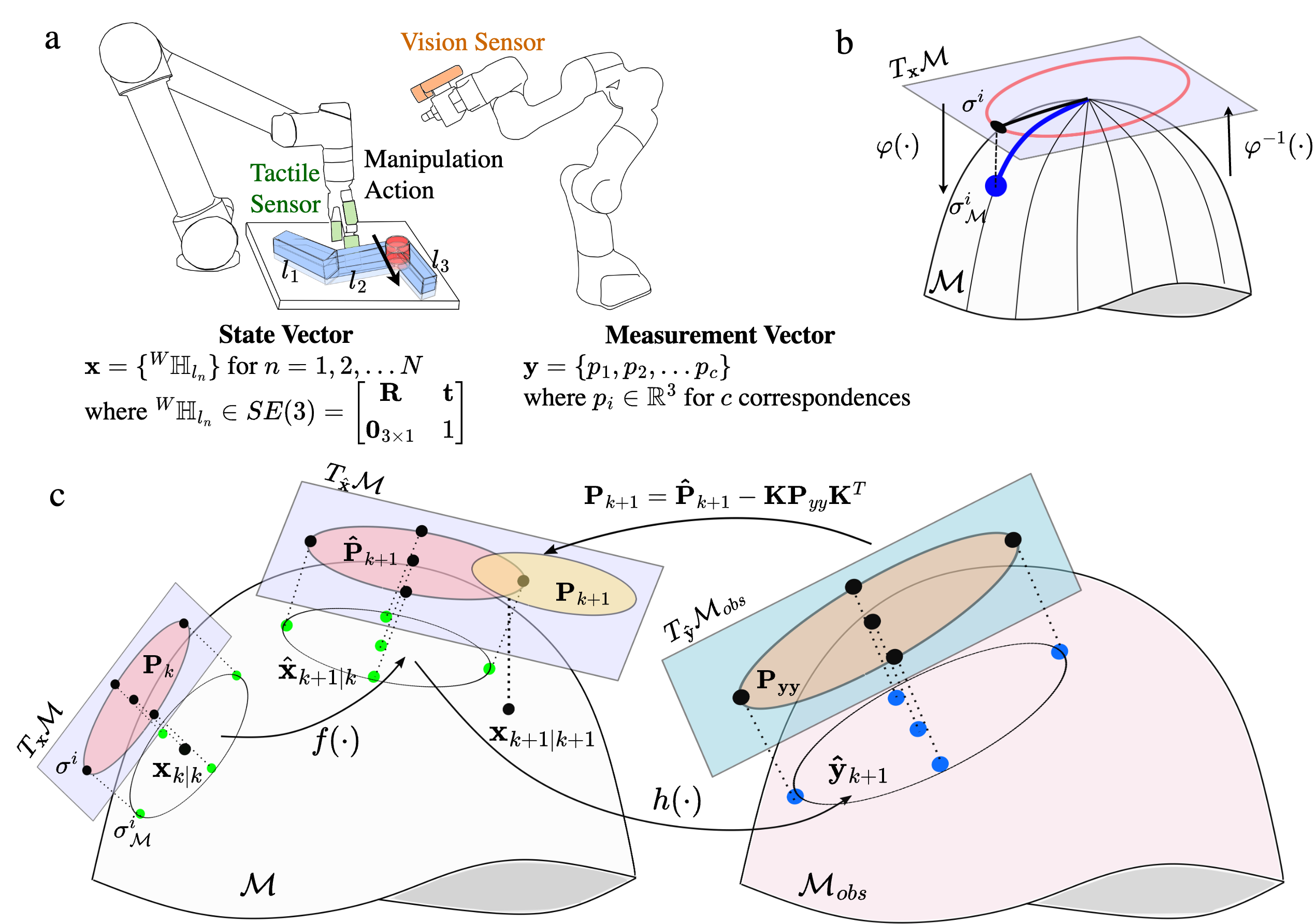}
    \caption{(a) The experimental setup shown along with the description of the state and measurement vector. (b) Visualization of the basic manifold operations: exponential map ($\varphi(\cdot)$) and logarithmic map ($\varphi^{-1}(\cdot)$). (c) Our ArtReg algorithm which is a manifold unscented Kalman filter visualised with operations on the state manifold $\mathcal{M}$ and measurement manifold $\mathcal{M}_{obs}$.}
    \label{fig:mukf}
\end{figure}

Let us consider a system with known initial mean and initial covariance as $\Bar{\mathbf{x}}_0 = E[\mathbf{x}_0]$ and ${\mathbf{P}}_0 = P[\mathbf{x}_0]$ respectively. 
In our case, the state $\hat{\mathbf{x}} \in SE(3)$ describes the 6 DoF pose of an object and evolves in the manifold $\mathcal{M}$ as described in Fig.~\ref{fig:mukf}.
We aim to obtain the \textit{a posteriori} state estimate $\hat{\mathbf{x}}_{k|k}$ and \textit{a posteriori} covariance $\hat{\mathbf{P}}_{k|k}$ upto time $k$ integrating $k$ observations. Given the state transitions model (also called process model) as $f(\cdot)$ and measurement model as $h(\cdot)$. The \textit{a posteriori} state can be obtained as follows: 
\begin{equation}
    \hat{\mathbf{x}}_{k|k} = \hat{\mathbf{x}}_{k|k-1} + \mathbf{K}(\mathbf{y}_{k|k} - \hat{\mathbf{y}}_{k|k-1})
    \label{eq:state_estimation}
\end{equation}
wherein, $\mathbf{K}$ is the Kalman gain, $\hat{p}$ refers to prediction of variable $p$, and $\mathbf{y}$ is the measurement observation. When $f(\cdot)$ and $h(\cdot)$ are non-linear functions, an Unscented Kalman Filter (UKF) can be used~\citep{wan2000unscented}. The UKF uses the \textit{Unscented Transformation} technique which picks a minimal set of sample points, called sigma points around the mean. These sigma points are propagated through the non-linear functions from which a new mean and covariance are obtained. Consider the state as $M$ dimensional, the sigma points $\sigma^i$ and the corresponding weights $W_i$ around the mean of the state are calculated as follows:
\begin{align}
\begin{split}
    \sigma^0 &= \bar{\mathbf{x}} \\
    \sigma^i &= \bar{\mathbf{x}} + (\sqrt{(M+\lambda)P})_i \quad i=1, \dots M \\
    \sigma^i &= \bar{\mathbf{x}} - (\sqrt{(M+\lambda)P})_{i-M} \quad i= M+1, \dots 2M \\
    W^{(m)}_0 &= \lambda / (M+\lambda) \\
    W^{(c)}_0 &= \lambda / (M+\lambda) + (1 - \alpha^2 + \beta) \\
    W^{(m)}_i &= W^{(c)}_i = 1/2(M+\lambda) \quad i = 1, \dots 2M
    \label{eq:sigma_pt_calculation}
\end{split}
\end{align}

where $\lambda = \{\alpha^2(M+\kappa) - M \}$, and $\alpha, \beta, \kappa$ are scaling parameters, $(\sqrt{(M+\lambda)P}_i)$ refers to $i^{th}$ column of the matrix square root.

The tangent space at $\mathbf{x}$ is defined as $T_{\mathbf{x}}\mathcal{M}$. There are two sets of sigma points that describe the covariance $\mathbf{P}$ in the manifold and the tangent space, i.e., $\sigma^{i} \in T_{\mathbf{x}}\mathcal{M}$ and $\sigma^{i}_{\mathcal{M}} \in \mathcal{M}$ wherein $i = 0,1, \dots, 2M$. The sigma points on the tangent (Euclidean) space are the same as described in Eq.~\eqref{eq:sigma_pt_calculation}. The mapping $\varphi(\cdot)$ also called as \textit{retraction} transforms a point in the tangent space into the manifold space and $\varphi^{-1}(\cdot)$ performs the inverse operation as described in Fig.~\ref{fig:mukf}b. In the SE(3) Lie Group, $\varphi(\cdot)$ is also called as the exponential map ($\mathtt{exp}$) that maps the elements of the Lie algebra to the Lie Group and $\varphi^{-1}(\cdot)$ is the logarithmic map ($\mathtt{Log}$).

\textbf{Prediction Step:} The mean is propagated using the non-noisy process model $f(\cdot): \mathcal{M} \rightarrow \mathcal{M}$ and input vector $u_k$ as:
\begin{equation}
    \hat{\mathbf{x}}_{k|k-1} = f(\hat{\mathbf{x}}_{k-1|k-1}, u_k, \mathbf{0})
    \label{eq:mukf-state}
\end{equation}

Contrary to the classical Euclidean UKF, wherein the weighted mean of the sigma points are propagated through the process model, in the manifold UKF, propagating the state directly through the process model avoids the computational complexity of finding the weighted averaging on manifold $\mathcal{M}$~\citep{brossard2020code}. To compute the covariance $\mathbf{P}_{k|k-1}$, the sigma points are propagated through the process model as:
\begin{equation}
    \mathbf{P}^s_{k|k-1} = \sum_{i=1}^{2M} W_i\varphi^{-1}_{\hat{\mathbf{x}}_{k|k-1}}(f(\sigma_{\mathcal{M}}^i))(\varphi^{-1}_{\hat{\mathbf{x}}_{k|k-1}}(f(\sigma_{\mathcal{M}}^i)))^T
    \label{eq:mukf-state_noise}
\end{equation}
Similarly, the noise sigma points are also propagated through the process model and added to the covariance as:
\begin{equation}
    \mathbf{P}^n_{k|k-1} = \sum_{i=1}^{2M} W_i\varphi^{-1}_{\hat{\mathbf{x}}_{k|k-1}}(f(\sigma_{\mathcal{M}}^{'i}))(\varphi^{-1}_{\hat{\mathbf{x}}_{k|k-1}}(f(\sigma_{\mathcal{M}}^{'i})))^T
    \label{eq:mukf_noise}
\end{equation}
wherein $\sigma^{'i} = \pm(\sqrt{(M+\lambda)\mathbf{Q}})_i$. The covariance is calculated as:
\begin{equation}
    \mathbf{P}_{k|k-1} = \mathbf{P}^s_{k|k-1} + \mathbf{P}^n_{k|k-1} 
    \label{eq:mulkf_covariance}
\end{equation}

\textbf{Update Step:} The measurement model $h(\cdot)$ generates a predicted observation for each sigma point. We note that the predicted observation $\hat{\mathbf{y}}_{k|k-1}$ belongs to the observation manifold $\mathcal{M}_{obs}$ such that $h(\cdot): \mathcal{M} \rightarrow \mathcal{M}_{obs}$. The predicted measurements are calculated using sigma points on the manifold:
\begin{equation}
    \hat{\mathbf{y}}_{k|k-1} = \sum_{i=0}^{2M} W_i^{(m)}h(\sigma^i_{\mathcal{M}}) 
    \label{eq:predicted_obs_manifold}
\end{equation}
The covariance and the cross-covariance are calculated similarly using the sigma points on the manifold as:
\begin{align}
    \begin{split}
        \mathbf{P}_{yy} &= \sum_{i=0}^{2M}W^{(c)}_i(h(\sigma^i_{\mathcal{M}}) - \hat{\mathbf{y}}_{k|k-1} )(h(\sigma^i_{\mathcal{M}}) - \hat{\mathbf{y}}_{k|k-1} )^T + \mathbf{R}_k \\
        \mathbf{P}_{xy} &= \sum_{i=1}^{2M}W^{(c)}_i\sigma^i_{\mathcal{M}}(h(\sigma)^i_{\mathcal{M}} - \hat{\mathbf{y}}_{k|k-1} )^T \quad .
    \end{split}
    \label{eq:covariance_manifold}
\end{align}

Once the predicted covariance and cross-covariance is calculated, the Kalman gain is computed by 
\begin{equation}
    \mathbf{K}_k =  \mathbf{P}_{xy} \mathbf{P}_{yy}^{-1} 
    \label{eq:kalman_gain}
\end{equation}

 Note that the covariance and cross-covariance lies on the tangent spaces $T_{\hat{\mathbf{x}}}\mathcal{M}$ and $T_{\hat{\mathbf{y}}}\mathcal{M}_{obs}$ respectively. The updated covariance is calculated identically using the Kalman gain as:
 \begin{align}
    \begin{split}
         \mathbf{P}_{k|k} &= \mathbf{P}_{k|k-1} - \mathbf{K}_k\mathbf{P}_{yy}\mathbf{K}_k^T
    \end{split}
    \label{eq:updated_state_cov}
\end{align}

The updated state is calculated using the retraction back onto the manifold as:
\begin{equation}
    \hat{\mathbf{x}}_{k|k} = \varphi(\hat{\mathbf{x}}_{k|k-1}, \mathbf{K}(\mathbf{y}_{k|k} - \hat{\mathbf{y}}_{k|k-1}))
    \label{eq:mukf_satte_update}
\end{equation}
The process is shown graphically in Fig.~\ref{fig:mukf}c. 

In the case of tracking objects, the state $\mathbf{x}$ describes the 6DoF pose of an object to be tracked and lies in $SE(3)$ Lie Group. In general, for any arbitrary $N$ objects that need to be tracked, we can define state variables $\mathbf{x}^{i}$ for $i=1,2, \dots, N$ which are tracked independently using $N$ manifold UKFs. The formulation is general and can be deployed for single, multiple, or articulated object tracking in an identical manner. Our framework for ArtReg is described in Fig.~\ref{fig:framework}a.
For a given point cloud of objects captured by the camera, the number of objects to be tracked is provided by the segmentation method described in Sec.~\ref{sec:part_segmentation}. If tactile contact points are available during robot interaction, these are added to the observed point cloud. For each object $O^{i=1:N}$, the segmented point cloud at the initial state at time $t=0$ corresponds to the object point cloud $\mathcal{O}^{i=1:N}$. At each time instant $k$, we seek to track the pose $\mathbf{x}^i_{k|k} \in SE(3)$ of $O_i$ given the previous state information $\mathbf{x}^i_{k|k-1}$ and measurements $\mathbf{y}^i_{k}$. For the sake of simplicity of notations, we will describe the pose tracking process for a single object $O$. Extending the procedure for $N$ objects is trivial as the states $\mathbf{x}^{i}$ are independent of each other for $i\in[1,N]$. 

The process model $f(\cdot)$ is defined as a random walk model as  
\begin{equation}
    \hat{\mathbf{x}}_{k|k-1} = \hat{\mathbf{x}}_{k-1|k-1} + w_k
\end{equation}
where the noise model $w_k \sim \mathcal{N}(0, \mathbf{Q})$ represents a zero-mean Gaussian noise. The state is time-invariant and the random walk is induced through the state noise.

The measurement model $h(\cdot)$ defines the expected measurement $\hat{\mathbf{y}}_{k|k-1}$ when the object $O$ is at the predicted state $\hat{\mathbf{x}}_{k|k-1}$. Let the point cloud extract at time step $k$ be denoted as $\mathcal{S}^{full}_k$. We perform instance segmentation to extract the various objects present in the scene $\mathcal{S}^{j}_k$ for $j=1:L$ where in ideal cases $L=N$ and in noisy conditions or close overlap between objects $L\neq N$.
The measurement vector consists of 3D points ($x,y,z$) belonging to object $\mathcal{S}^j$ i.e., $\mathbf{y}_{k|k} = \{p_1, p_2 \dots p_Q\} \forall p \in \mathcal{S}^j_k$.  
The correspondence search finds the closest corresponding points in $\mathcal{O}$ and $\mathcal{S}^j$ i.e.,
\begin{equation}
    \xi(\sigma_\mathcal{M}) = argmin_{p \in \mathcal{O}}(||\mathcal{S}^j - \sigma_\mathcal{M}p ||)
    \label{eq:measurement_model_artreg}
\end{equation}
The 3D points belong to the point cloud of the object $\mathcal{O}$ are transformed to the predicted pose for each sigma point $\sigma_{\mathcal{M}} \in SE(3)$. 
Since there could be multiple objects in the scene such that $\mathcal{S}^{j=1:L}$, the correspondence search ensures to find the object with the closest Euclidean distance compared with the predicted pose of the object.
The measurement model provides the correspondence points in the object point cloud with respect to the measured scene point cloud such that $h(\sigma_{\mathcal{M}}^{(m=0:2M)}) = \xi(\sigma_{\mathcal{M}}^{(m=0:2M)})$.
The \textit{a posteriori} state is obtained by minimizing the distance between the correspondence points between the object point cloud and the scene point cloud ($\mathbf{y}_{k|k} - \hat{\mathbf{y}}_{k|k-1}$) using Eq.~\ref{eq:mukf_satte_update}. The state and measurement noise ($\mathbf{Q}$ and $\mathbf{R}$ respectively) are initialized as diagonal matrices with random values. The ratio between the state and measurement noise determines if our filter tracks the predicted states (if state noise is lower) or if it tracks the measurements (if the measurement noise is lower). Since our process models resembles a random walk, we initialize the measurement noise lower than the state noise in order to track the dynamically moving object(s). The overall algorithm is shown in Algorithm~\ref{algo}.

\begin{algorithm}[]
\SetAlgoLined
\KwIn{ $\hat{\mathbf{x}}_{k-1|k-1}, \mathbf{P}_{k-1|k-1}, \mathbf{R}_{k-1}, \mathbf{Q}_{k-1}, u_k$} 
\KwOut{$\hat{\mathbf{x}}_{k|k}, \mathbf{P}_{k|k}, \mathbf{K}_k$}
\textbf{Prediction:} \\
\CommentSty{//Propagate the mean state}\\
$\hat{\mathbf{x}}_{k|k-1} = f(\hat{\mathbf{x}}_{k-1|k-1}, u_k, \mathbf{0})$ \\
\CommentSty{//Compute the sigma points on tangent space and manifold}\\
$\sigma^i, \sigma_{\mathcal{M}}^i$ for $i=0,\dots,2M$ where $M=6$. \Comment*[r]{Eq. \eqref{eq:sigma_pt_calculation}} 
\CommentSty{//State Covariance and Noise Propagation} \\
$ \mathbf{P}^s_{k|k-1} = \sum_{i=1}^{2M} W_i\varphi^{-1}_{\hat{\mathbf{x}}_{k|k-1}}(f(\sigma_{\mathcal{M}}^i))(\varphi^{-1}_{\hat{\mathbf{x}}_{k|k-1}}(f(\sigma_{\mathcal{M}}^i)))^T$ \\
  $\mathbf{P}^n_{k|k-1} = \sum_{i=1}^{2M} W_i\varphi^{-1}_{\hat{\mathbf{x}}_{k|k-1}}(f(\sigma_{\mathcal{M}}^{'i}))(\varphi^{-1}_{\hat{\mathbf{x}}_{k|k-1}}(f(\sigma_{\mathcal{M}}^{'i})))^T$\\
   $\mathbf{P}_{k|k-1} = \mathbf{P}^s_{k|k-1} + \mathbf{P}^n_{k|k-1}$ \\ 
\textbf{Update:} \\
$\mathbf{y}_{k|k}$ \Comment*[r]{Get actual measurement from robot} 
\CommentSty{//Compute measurement sigma points}\\
$ \hat{\mathbf{y}}_{k|k-1} = \sum_{i=0}^{2M} W_i^{(m)}h(\sigma^i_{\mathcal{M}}) $ \\
\CommentSty{//Compute covariance and cross-covariance matrix}\\
$\mathbf{P}_{yy} = \sum_{i=0}^{2M}W^{(c)}_i(h(\sigma^i_{\mathcal{M}}) - \hat{\mathbf{y}}_{k|k-1} )(h(\sigma^i_{\mathcal{M}}) - \hat{\mathbf{y}}_{k|k-1} )^T + \mathbf{R}_k$ \\
$\mathbf{P}_{xy} = \sum_{i=1}^{2M}W^{(c)}_i\sigma^i_{\mathcal{M}}(h(\sigma)^i_{\mathcal{M}} - \hat{\mathbf{y}}_{k|k-1} )^T$\\
\CommentSty{//Compute Kalman Gain}\\
$\mathbf{K}_k =  \mathbf{P}_{xy} \mathbf{P}_{yy}^{-1} $\\
\CommentSty{//Compute Mean and Covariance update}\\
$\mathbf{P}_{k|k} = \mathbf{P}_{k|k-1} - \mathbf{K}_k\mathbf{P}_{yy}\mathbf{K}_k^T$ \\
$\hat{\mathbf{x}}_{k|k} = \varphi(\hat{\mathbf{x}}_{k|k-1}, \mathbf{K}(\mathbf{y}_{k|k} - \hat{\mathbf{y}}_{k|k-1}))$
\caption{ArtReg: Object tracking with Manifold Unscented Kalman Filter}
\label{algo}
\end{algorithm}

\subsection{Visuo-Tactile Interactive Perception for Articulation Detection}
\label{sec:articulation_deetction}
Interactive perception techniques and our pose tracker, ArtReg are used together to distinguish articulated objects from rigid objects as well as identifying the type of articulation joints present. In this work, we limit ourselves to objects with revolute and prismatic joints of 1 degree of freedom (DoF). However, we can trivially extend our method for n-DoF systems. The visual point clouds of the objects are segmented using our instance segmentation approach to obtain the estimated number of parts present and possible articulation joint (object kinematic model). We develop an iterative visuo-tactile based interactive perception approach which updates the belief regarding the object kinematic model with each manipulation action. These actions which are \textit{push} or \textit{hold-pull} are chosen and performed autonomously by the robot. The actions are chosen in greedy manner which allows extracting the correct object kinematic structure with minimal number of actions. We detail our instance segmentation approach, action selection and execution below.

\subsubsection{Part Segmentation}
\label{sec:part_segmentation}
We consider objects placed on a planar surface which needs to be segmented from the background and separated into different parts depending on the object configuration. From the RGB point cloud of the scene, firstly the background plane is segmented using RANSAC-based plane fitting method wherein all points that lie within a user-defined distance threshold of a best-fitting plane are filtered out. The plane parameters are extracted in the form $ax+by+cz+d=0$ format and the points that lie within a specified distance threshold to the plane are regarded as inliers and used for segmentation.
Secondly, during interactive perception the robot is also in contact with the object of interest and possibly occludes the object. Hence, it is necessary to also segment the robot links from the RGB point cloud. Given the current joint pose information of the robot body which are obtained from the robot controller, the robot mesh is projected onto the point cloud and the 3D points which correspond to the robot body are filtered out from the point cloud. Thirdly, the remaining point cloud consists of the objects and possibly noisy outlier points which are filtered out using statistical outlier removal methods that removes points further away from neighborhood points compared to the average for the point cloud. Finally, there may be multiple objects present in the scene with optional kinematic linkages between them. Hence to perform part segmentation, we used a region-growing segmentation algorithm~\citep{rusu2008functional}. Given an initial set of seed points, each neighborhood point is checked if it belongs to the same region as the seed point or if it should be considered as a future seed point in a future iteration. The criteria for inclusion in the same region depends on a smoothness constraint (based on surface normals) and colour information. After an initial segmentation process, regions having similar colour information are merged together. We used an implementation of the region growing algorithm available in the Point Cloud Library (PCL)~\citep{rusu20113d}. As described in the framework in Fig.~\ref{fig:framework}b, after each action is performed, the part segmentation provides the belief of the number of objects in the scene which is updated upon subsequent actions. An object is considered fully explored if the belief regarding its kinematic model remains consistent over three consecutive iterations.

\begin{figure}[th!]
    \centering
    \includegraphics[width = 0.9\textwidth]{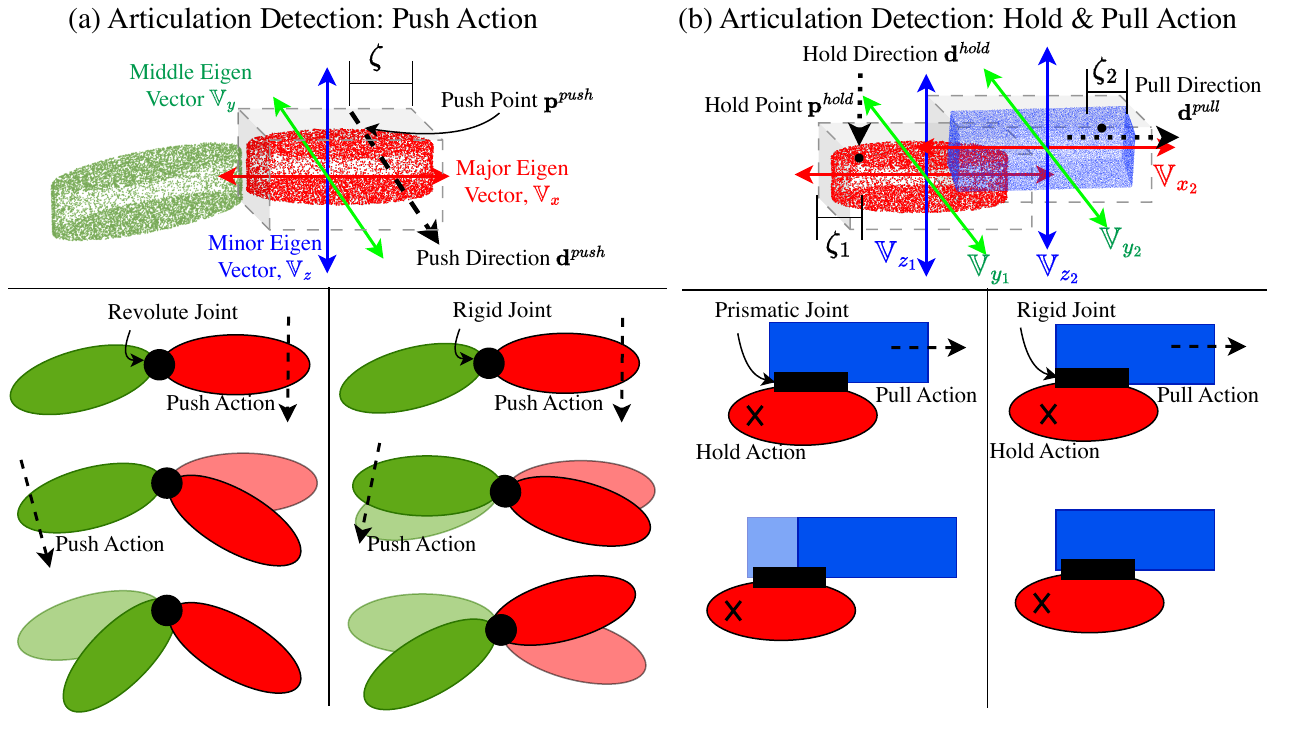}
    \caption{Interactive perception for articulation detection: (a) push action for revolute joints, (b) grasp and pull action for prismatic joints.}
    \label{fig:action_exec}
\end{figure}
\subsubsection{Interactive Perception Action Selection} We define two types of actions: \textit{push} and \textit{hold-pull} actions. The push actions are used to distinguish if an object is rigid or articulated with a revolute joint. The hold-pull action is used only to distinguish if an object has a prismatic joint. Our object exploration strategy is as follows: if the current belief from the part segmentation has multiple objects, then the robot always begins with pushing actions. The object(s) are tracked, and the change in relative pose after performing the action is used to infer the joint type. The heuristic for joint-type inference is as follows: if the object pushes one part and there is a change in pose registered for both parts, then it is likely not to have a revolute joint. In that case, the robot proceeds to push the other part and if a similar behavior is observed, then the belief is confirmed to be either a rigid object or an object with prismatic joint. In contrast, if only the pushed part is moved while the remaining parts are static, it is likely that the parts are connected to a revolute joint. The robot pushes the other part consequently to confirm the object kinematic structure. This process can be seen in Fig.~\ref{fig:action_exec}a. If pushing action is not possible to distinguish the object kinematics conclusively, the system reverts to performing a hold-pull action sequence. One robot holds one part by applying an orthogonal force to the surface of the object and another robot tries to grasp and pull the other part. If a change in pose is detected after performing the action, it can be concluded that the parts are linked through a prismatic joint. This process can be seen in Fig.~\ref{fig:action_exec}b.

\subsubsection{Action Execution}
\label{sssec:action_exec}
For both types of the actions i.e., push and hold-pull, the action parameters are based on geometric information extracted from the point cloud. The oriented bounding box (OBB) of the point cloud representing the object is extracted. 
We extract the Eigen vectors from the covariance matrix of the point cloud representing the object. The major Eigen vector $\mathbb{V}_x$ represents the axis along the length of the object, the middle Eigen vector $\mathbb{V}_y$ is perpendicular to the object and the minor Eigen vector $\mathbb{V}_z$ is normal to the surface of the object. Push action is parameterized by a push position and direction, $a^{push} = \{\mathbf{p}^{push}, \mathbf{d}^{push}\}$. For detecting articulated objects, the push direction is parallel to $\mathbb{V}_y$ and the push position is along the face of the oriented bounding box that is parallel to $\mathbb{V}_y$ with a tolerance distance $\zeta$ such that the push direction intersects along the edge of the object.
The pushing distance is predefined for 10 cm for articulation detection. The push action is described in Fig.~\ref{fig:action_exec}a.

In contrast, for the hold-pull action, both robots are used similar to how humans perform bimanual manipulation actions to pull apart two parts linked by a prismatic joint. Referring to Fig.~\ref{fig:fig1}, the Panda robot which is equipped with the RGB-D vision camera but without tactile sensing is used for performing the hold action. The Panda robot also has force/torque sensors on all joints which are used for proprioceptive information.
The UR5 robot which is sensorised with tactile sensing is used to perform the grasp and pull action. In order to ensure that the robots do not collide with each other, the actions are performed close to the diametrically opposing edges of the object. The hold and pull actions are performed on the two different parts of the object.
The hold action is parameterized by the hold position and direction as $a^{hold} = \{\mathbf{p}^{hold}, \mathbf{d}^{hold}\}$. Given two parts of the object denoted by 1 and 2, respectively, the hold direction $\mathbf{d}^{hold}$ is parallel to the minor Eigen vector for the one of the object parts as $\mathbb{V}_{z_1} \ ||\ \mathbf{d}^{hold}$. The hold position is obtained by moving the centroid of the part point cloud to the edge of the bounding box with a certain tolerance distance $\zeta$ from the bounding box edge. The force-torque sensor on the joint 7 of the Panda robot is used to detect contact and apply a constant holding force on the object part.
Similarly, the pull action is parameterized by position and direction $a^{pull} = \{\mathbf{p}^{pull}, \mathbf{d}^{pull}\}$. In order to grasp objects of different shapes, when provided with a position $\mathbf{p}^{pull}$, the robot is moved to a position vertically above the $\mathbf{p}^{pull}$ at a predefined height and the gripper opens to the maximum width. The robot moves in a straight-line downwards to the $\mathbf{p}^{pull}$ and closes the gripper until contact is detected by the tactile sensors. The gripper applies a constant grasping force on the object part. The position $\mathbf{p}^{pull}$ is obtained by shifting the centroid of the object part $2$ along $\mathbb{V}_{x_2}$ to the edge of the bounding box with a certain tolerance distance $\zeta$ from the edge. The pull direction is parallel to the major Eigen vector of the other object part such that  $\mathbb{V}_{x_2} \ || \ \mathbf{d}^{pull}$ .  
Once the robot grasps the object part, it pulls for a predefined distance of 5 cm. The tactile sensors on the inner side of the gripper fingers are used to detect possible loss of contact and stop the robot. The hold-pull maneuver is described in Fig.~\ref{fig:action_exec}b.

\subsection{Visuo-Tactile Goal-driven Closed-Loop Control}
\label{sec:goal-driven-pushing}
\begin{figure}
    \centering
    \includegraphics[width = 0.7\textwidth]{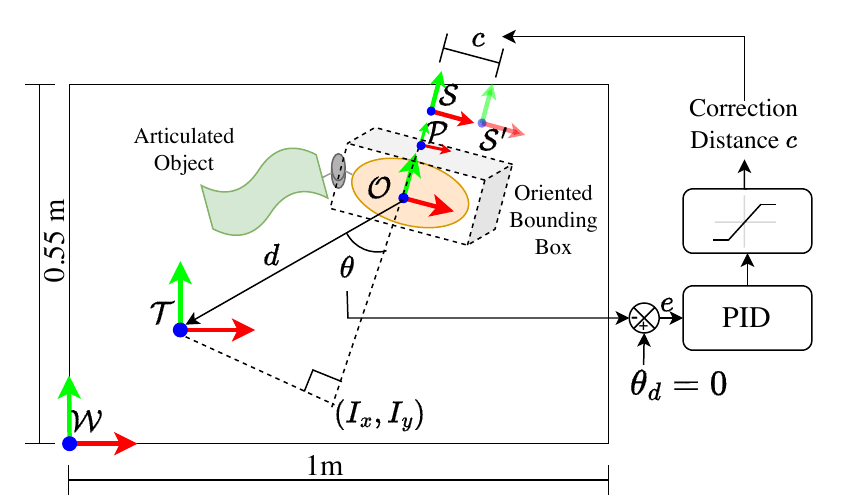}
    \caption{Goal-driven closed loop control system}
    \label{fig:controller}
\end{figure}
The objective of the closed-loop controller is to push the object (single or articulated) to a pre-defined goal pose. While the closed-loop controller is identical for both single or articulated objects, in the case of articulated objects a task planner decides which part of the articulated object to push based on the distance to goal. Without loss of generality, we detail the push controller for a single object as shown in Fig.~\ref{fig:controller}. 
Definition of the coordinate frames used to devise the controller: the geometric center of the oriented bounding box (OBB) is defined as $\mathcal{O}$; the target frame $\mathcal{T}$; pushing frame as $\mathcal{P}$; tactile sensor frame as $\mathcal{S}$ and world coordinate frame as $\mathcal{W}$. The objective is to align $\mathcal{O}$ with $\mathcal{T}$ within a defined tolerance error $\lambda$ such that $d\rightarrow0, \theta\rightarrow0$. Given the current pose of the object as ${}^\mathbf{W} H_{\mathcal{O}}$, the push pose ${}^\mathbf{W} H_{\mathcal{P}}$ is computed along the middle Eigen vector $\mathtt{V}_y$ such that it passes through the geometric center $\mathcal{O}$ as detailed in Sec.~\ref{sssec:action_exec}. However, when pushing to a predefined goal, the push position $\mathbf{p}^{push}$ is shifted along the perimeter of the object such that the bearing (angular offset from the goal) is minimised. The push direction vector $\mathbf{d}^{push}$ is kept the same in order to ensure the maximum area of the tactile sensor array is in contact with the object. To minimise $\theta$, the sensor frame is shifted along the object perimeter to $\mathbf{S}'$ based on the output from a proportional-integral-derivative (PID) controller. The angular offset $\theta$ is measured as follows: The slope of the line $l_{\mathcal{P}\mathcal{O}}$ passing through $(\mathcal{O}_x, \mathcal{O}_y)$ and $(\mathcal{P}_x, \mathcal{P}_y)$ is 
\begin{equation}
    m = \frac{\mathcal{P}_y - \mathcal{O}_y}{\mathcal{P}_x - \mathcal{O}_x} \quad .
\end{equation}
The perpendicular projection of the point ($\mathcal{T}_x, \mathcal{T}_y$) on the line $l_{\mathcal{P}\mathcal{O}}$ has a negative reciprocal slope $-1/m$. The equation of perpendicular projection from $\mathcal{T}$ is $(y-\mathcal{T}_y) = \frac{-1}{m}(x- \mathcal{T}_x)$. Solving for the point of intersection ($I_x, I_y$) we get:
\begin{align}
\begin{split}
    I_x &= \frac{\mathcal{T}_x - m^2\mathcal{O}_x - m(\mathcal{O}_y - \mathcal{T}_y)}{1+ m^2} \\
    I_y &= (-1/m)I_x + \mathcal{T}_y
\end{split}
\end{align}
The angle $\theta$ is computed as:
\begin{equation}
    \theta = \tan^{-1}\left(\frac{\sqrt{(\mathcal{T}_x - I_x)^2 + \mathcal{T}_y - I_y)^2}}{\sqrt{(\mathcal{O}_x - I_x)^2 + \mathcal{O}_y - I_y)^2}}\right)
\end{equation}
The input to the PID control is $e = \theta - \theta_d$ where $\theta_d = 0$. The PID control output at time $t$ is provided as:
\begin{equation}
    c(t) = k_pe(t)+ k_i\int_0^te(\tau)d\tau + k_d\frac{de}{dt} \quad .
    \label{eq:pid}
\end{equation}
The PID gain coefficients $k_p, k_i, k_d$ were computed empirically through trail-and-error as 0.05, 0.0, 0.03 respectively. Ad-hoc tuning of the controller was performed due to the relatively simple control problem.
The controller output is passed through a saturation function wherein the output value $c$ is clipped if it is beyond $\pm3$ cm. The correction distance $c$ is incorporated into the sensor pose as:
\begin{equation}
    {}^{\mathcal{W}}H_{\mathcal{S}'} = {}^{\mathcal{W}}H_{\mathcal{S}}{}^{\mathcal{S}}H_{\mathcal{S}'}
\end{equation}
wherein ${}^{\mathcal{S}}H_{\mathcal{S}'} = \begin{bmatrix} 
 &  &  & c \\
 & \mathbb{I}_{3} &  & 0 \\
 &  &  & 0 \\
 & \mathbf{0}_{3x1} &  & 1 \\
\end{bmatrix}$. The distance to goal-pose is given as:
\begin{equation}
    d = \sqrt{(\mathcal{T}_x - \mathcal{O}_x)^2 + (\mathcal{T}_y - \mathcal{O}_y)^2} \quad .
    \label{eq:disttogoal}
\end{equation}
The pushing interaction is performed as a discrete iterative process wherein the robot performs the push action with 3 cm pushing distance. Tactile sensors are continuously monitored throughout the interaction to identify any instances of contact loss with the object. As seen from Fig.~\ref{fig:framework}c, the process of segmentation, pose estimation, push point and direction estimation and push execution are repeated until the object is aligned with the target pose with the user-specified tolerance bounds. A task planner is designed that coordinates the push actions and providing commands to the robot. For articulated objects, the iterative pushing strategy allows the robot to transition between different parts of the articulated object such that the distance to goal for each part is minimised uniformly. This ensures that the robot alternates to push each part of the articulated object iteratively. The tolerance to the goal-pose is set to 2 cm.

\section{Experimental Results}
\label{sec:experiments}
\subsection{Experimental Setup and Outline}
The experimental setup shown in Fig.~\ref{fig:fig1} consists of a Universal Robots UR5 with the Robotiq gripper that is sensorised with tactile sensor arrays and a Franka Emika Panda robot equipped with a RGB-D Azure Kinect DK vision sensor attached with a custom flange to the end-effector. The camera is used in an eye-in-hand configuration, which allows the Panda robot to actively perceive various parts of the objects in case of occlusion by the UR5 robot.
The tactile sensor arrays are obtained from XELA Robotics\footnote{\url{https://www.xelarobotics.com/}} and Contactile\footnote{\url{https://contactile.com/}}. Both these tactile sensor arrays are sensorised on each finger of the Robotiq 2F140 gripper and each sensing point of the arrays provides 3-axis force-torque measurements. We intentionally chose two different tactile sensors that have different operational principles to demonstrate that our method is agnostic to any particular type of tactile sensor. 
While our proposed method ArtReg is agnostic to the number of robots used, having two robots can help to manipulate certain objects, for instance, with prismatic joints where one robot can immobilise the object and the other robot can manipulate the movable link. 
All manipulation operations are performed within the robot workspace measuring 1.0 m × 0.55 m. The workspace limits also roughly match the kinematic limits of the UR5 robot which has a maximum kinematic reach of 0.85 m.

All experiments were performed on an Ubuntu 18.04 workstation with Intel Xeon Gold 5222 CPU. We used ROS Melodic as the robotic middleware, Point cloud library~\citep{rusu20113d} for operations involving point clouds, \texttt{kalmanif} library~\citep{Deray-20-JOSS} for Kalman filter implementations.

\begin{figure}
    \centering
    \includegraphics[width = 0.95\textwidth]{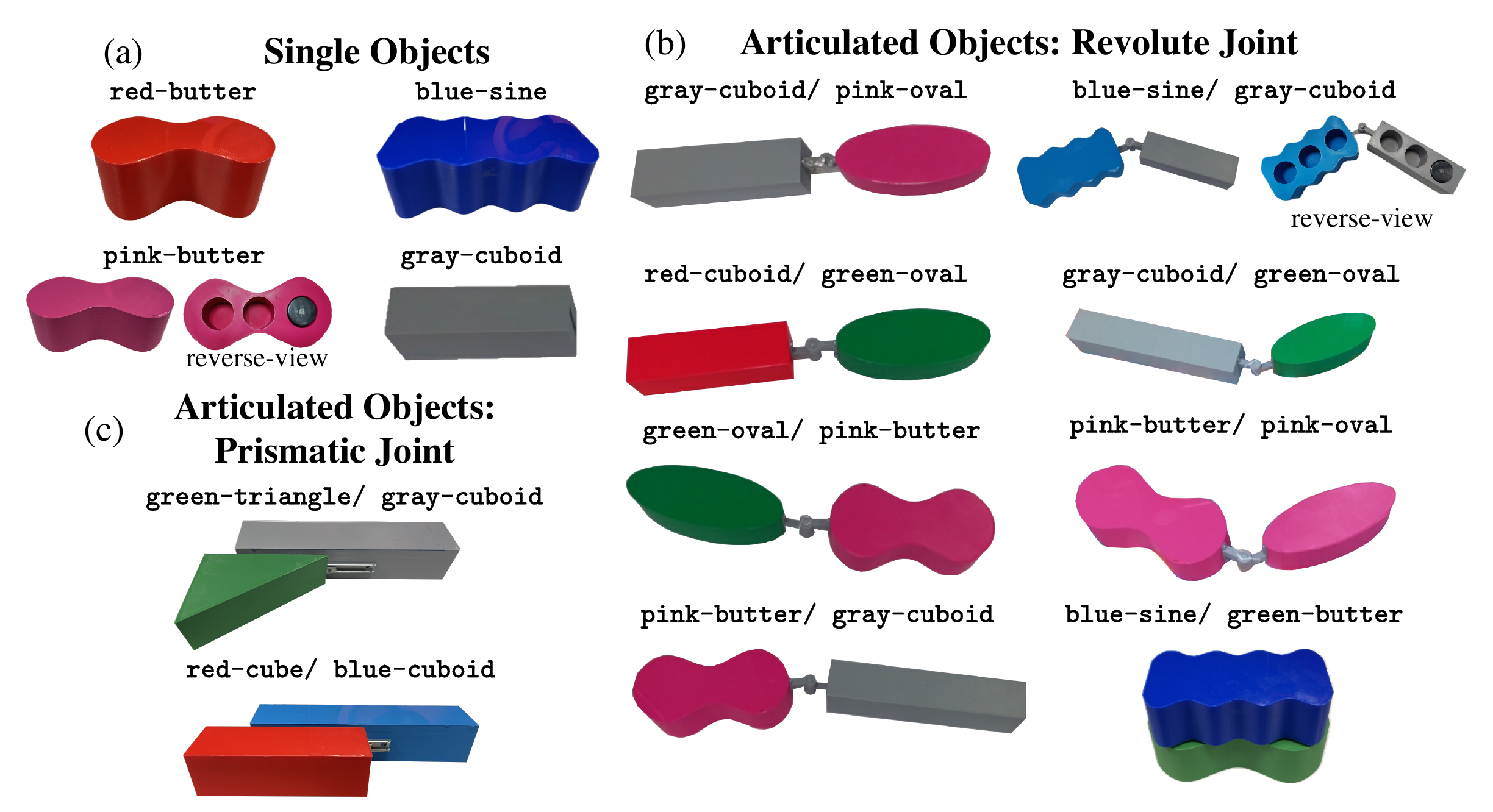}
    \caption{List of experimental objects used for tracking and closed-loop control: (a) single objects, (b) articulated objects with revolute joint and (c) articulated objects with prismatic joint. The reverse view of objects such as \texttt{blue-sine/ gray cuboid} and \texttt{pink-butter} is shown wherein the center of mass can be changed by placing an additional weight.}
    \label{fig:object-list}
\end{figure}

\textbf{Experimental Objects:} The experimental objects are shown in Fig.~\ref{fig:object-list}. The naming convention used is the \texttt{colour-shape} of the object. The objects have been 3D printed with varying sizes and shapes. We used the following shapes: \texttt{cube}, \texttt{cuboid}, \texttt{oval}, \texttt{butter}, \texttt{sine}, and \texttt{triangle}. The reverse side of all objects have 3 hollowed out cylinders (as seen in Fig.~\ref{fig:object-list}a - \texttt{pink-butter} and \texttt{blue-sine/ gray-cuboid}) in which external weights of 500g can be placed to offset the center of mass (CoM) of the objects. The external weight can be placed in any of the 3 possible holes unknown to the robotic system by the user and results in 3 different variations for CoM for each object.
The camera always sees the smooth `top' surface of the objects, hence the determination of offset in CoM can be performed using tactile sensing alone. Each of the objects can be attached to another with a revolute joint using a specialized 3D printed hinge (Fig.~\ref{fig:object-list}b). Furthermore, we used prismatic joints (typically used in drawers) attached to the objects axially to form an articulated prismatic object  (Fig.~\ref{fig:object-list}c). As a special case of the revolute joint shown as \texttt{blue-sine/ green-butter}, the object has an overlapping revolute joint such that the bottom object (\texttt{green-butter}) has a cylindrical protrusion and the top object (\texttt{blue-sine}) has the corresponding cylindrical hole allowing for a rotational joint of 1 DoF.

\textbf{Outline of experiments:}
We conducted a comprehensive and exhaustive set of robotic experiments to evaluate the entire pipeline and each experiment has been documented in the supplementary video. Detection of possible articulation as well as the type of articulated joints (revolute / prismatic) has been performed in Sec.~\ref{ssec:artDetect}. Closed loop control for goal-driven manipulation of singular and articulated objects are performed in Sec.~\ref{ssec:goalManip}. Furthermore, to evaluate for robustness we performed the goal-driven manipulation under conditions such as (a) low lighting, (b) challenging backgrounds, and (c) varying center-of-mass. In Sec.~\ref{ssec:tracking_exp} we performed only pose tracking without any target-driven robot manipulation for single, multiple and articulated objects. The number of repeated trials for each different experiment is detailed in the respective subsections.
The comparison with state-of-the-art approaches in demonstrated in Sec.~\ref{ssec:comps_exp} using a standard benchmark of synthetic articulated objects from the PartNet-Mobility dataset~\citep{Xiang_2020_SAPIEN}.

\subsection{Articulation Detection}
\label{ssec:artDetect}
\begin{figure}
    \centering
    \includegraphics[width = 0.9\textwidth]{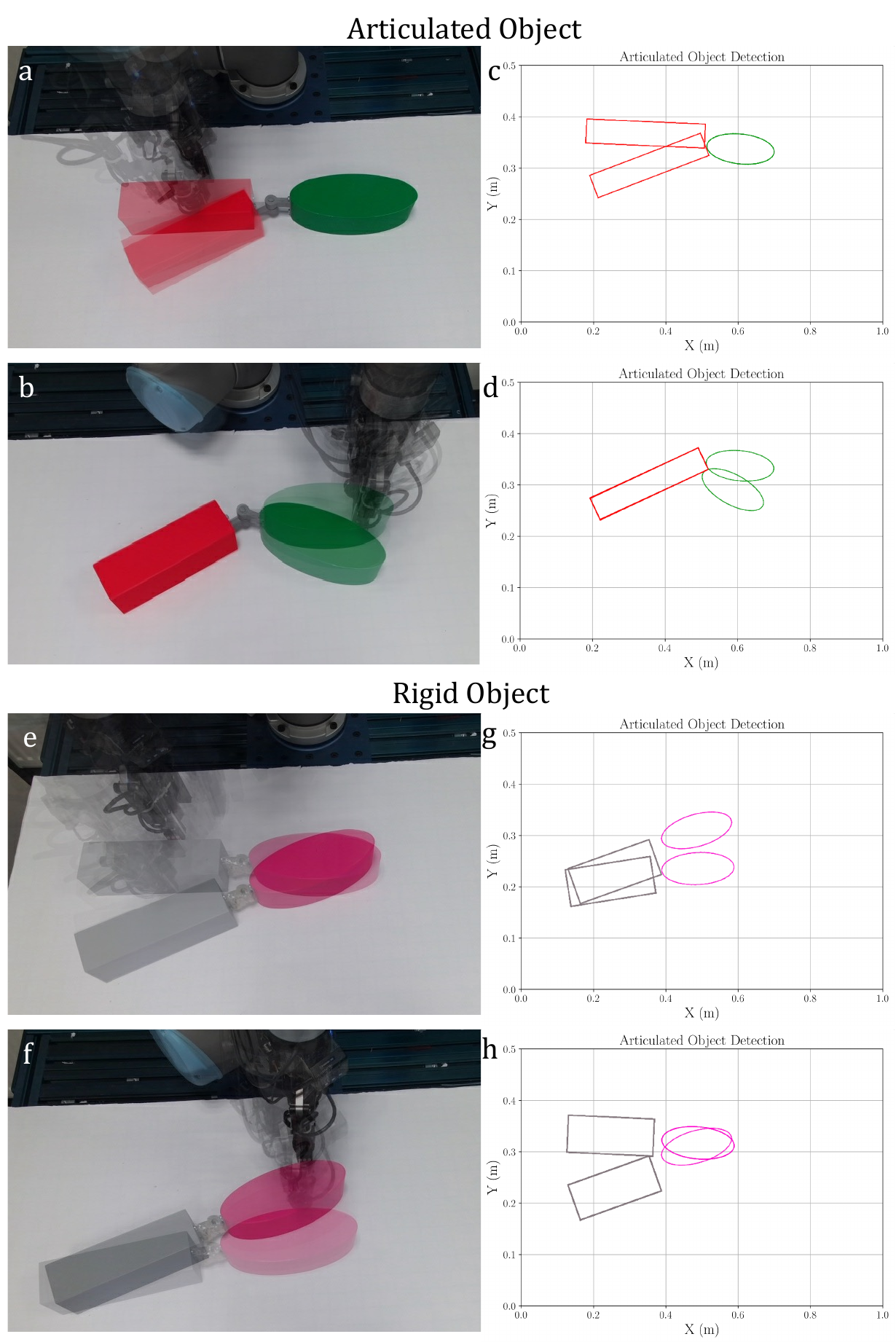}
    \caption{Detection of articulated revolute object with interactive perception (best viewed on screen and in colour)}
    \label{fig:rev_action_detection}
\end{figure}

For the detection of articulated objects from rigid objects, we used manipulation actions such as pushing and hold-pulling for interactive perception. In Fig.~\ref{fig:rev_action_detection}, two identical scenes are used wherein an \texttt{oval} object and \texttt{cuboid} object are connected with a revolute joint (Fig.~\ref{fig:rev_action_detection}a-d) and a rigid joint (Fig.~\ref{fig:rev_action_detection}e-h). Since visual inspection alone cannot differentiate between the rigid object and the articulated object, interactive perception is employed. In Fig.~\ref{fig:rev_action_detection}a, the robot pushes the \texttt{cuboid} for a predefined distance of 10 cm and our ArtReg tracker shows that only the \texttt{cuboid} is displaced as evidenced in Fig.~\ref{fig:rev_action_detection}b. Subsequently, the robot pushed the \texttt{oval} object, resulting in its displacement while the connected \texttt{cuboid} remained stationary. This observation led to the conclusion that the object is articulated. On the contrary, in Fig~\ref{fig:rev_action_detection}e, the robot pushed the \texttt{oval} object for 10cm and the \texttt{cuboid} object is also displaced. Similarly, the robot autonomously chooses to push the \texttt{cuboid} object next which also moves the \texttt{oval} object proportionally. This shows that the objects are connected by a rigid joint and infact is one rigid object.

\begin{figure}[b!]
    \centering
    \includegraphics[width = 0.95\textwidth]{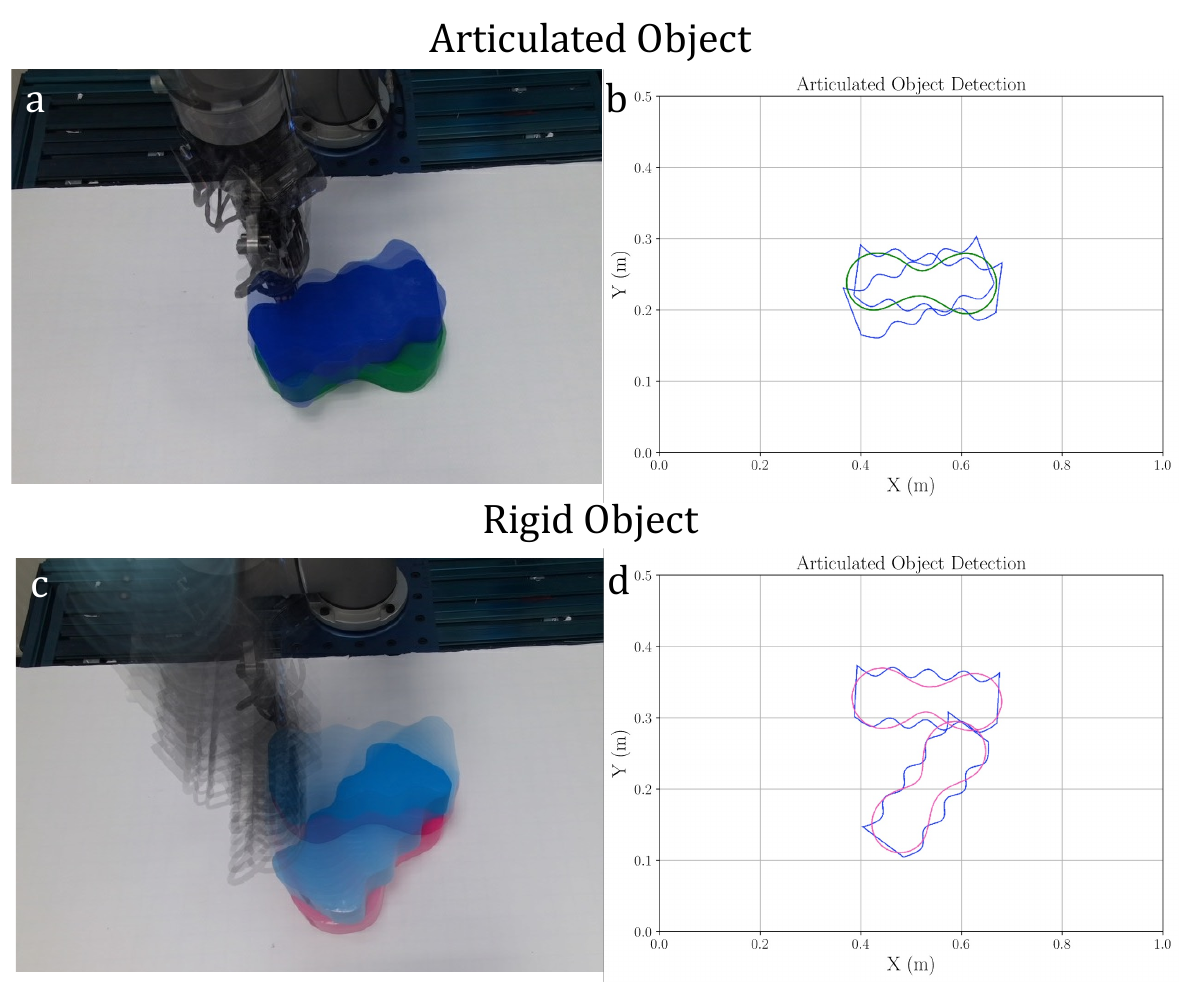}
    \caption{Detection of articulated object with overlapping revolute joint using interactive perception (best viewed on screen and in colour)}
    \label{fig:rev_top_action_detection}
\end{figure}

A similar experiment is performed with a configuration of objects with one object on top of another, connected together by a revolute joint and a rigid joint, respectively, as shown in Fig.~\ref{fig:rev_top_action_detection}. The robot used push action to distinguish an articulated object from the rigid object by tracking the motion of the two objects upon performing the action (see Fig.~\ref{fig:rev_top_action_detection}a,c for articulated object and Fig.~\ref{fig:rev_top_action_detection}b,d for rigid object).
Two identical objects, one with a rigid joint and one with an overlapping articulated joint are used. The robot uses the interactive push action to determining the object type. The push point is determined in a similar way as described in Fig.~\ref{fig:action_exec}a with the push point selected to the top (visible) surface of the object. The push action would result in a motion of only the top part of the object in case of an articulated joint (Fig.~\ref{fig:rev_top_action_detection}a-b) whereas the whole object is detected to move upon the interactive manipulation as seen in Fig.~\ref{fig:rev_top_action_detection}c-d.

\begin{figure}
    \centering
    \includegraphics[width = 0.95\textwidth]{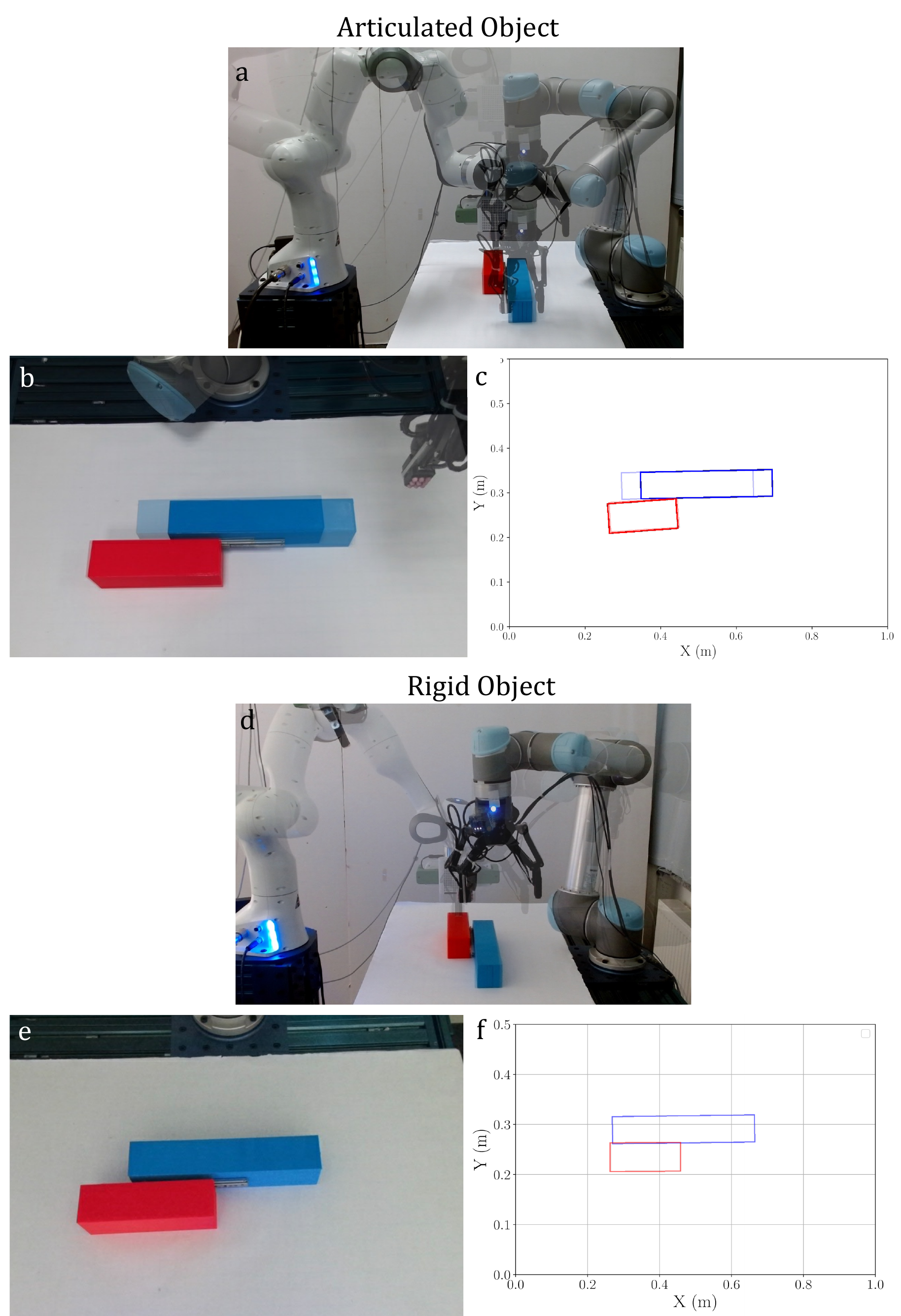}
    \caption{Detection of articulated prismatic object with interactive perception (best viewed on screen and in colour)}
    \label{fig:prism_action_detection}
\end{figure}

While pushing actions are sufficient to detect articulated objects with revolute joints, they cannot be used to distinguish objects with prismatic joints. 
Both robots are used to perform hold-pull maneuvers for such prismatic objects. In a similar experiment to the revolute joint, we used two sets of objects, a \texttt{red-cube} and \texttt{blue-cuboid} which are joint together with a prismatic joint and a rigid joint respectively. The Panda robot performs a hold maneuver on one of the object and the UR5 robot detects a possible pulling position by grasping the other object and pulling it for a fixed distance of 5 cm. The Panda robot used the force-torque sensor embedded at the end-effector joint to measure the constant holding force of around 8 N which is sufficient to immobilize the object without damaging it. Similarly, tactile sensors on the inside of the gripper finger pads are used to detect contact while grasping and pulling. Our ArtReg tracker is used to detect the change in pose of the objects which is used to distinguish the articulated object from the rigid object with prismatic joint as seen in Fig.~\ref{fig:prism_action_detection}. We performed five repeated trials for each case: revolute joint, top-revolute joint and prismatic joint with the corresponding object with rigid joints for comparison.

\begin{figure}[t!]
    \centering
    \includegraphics[width = 0.7\textwidth]{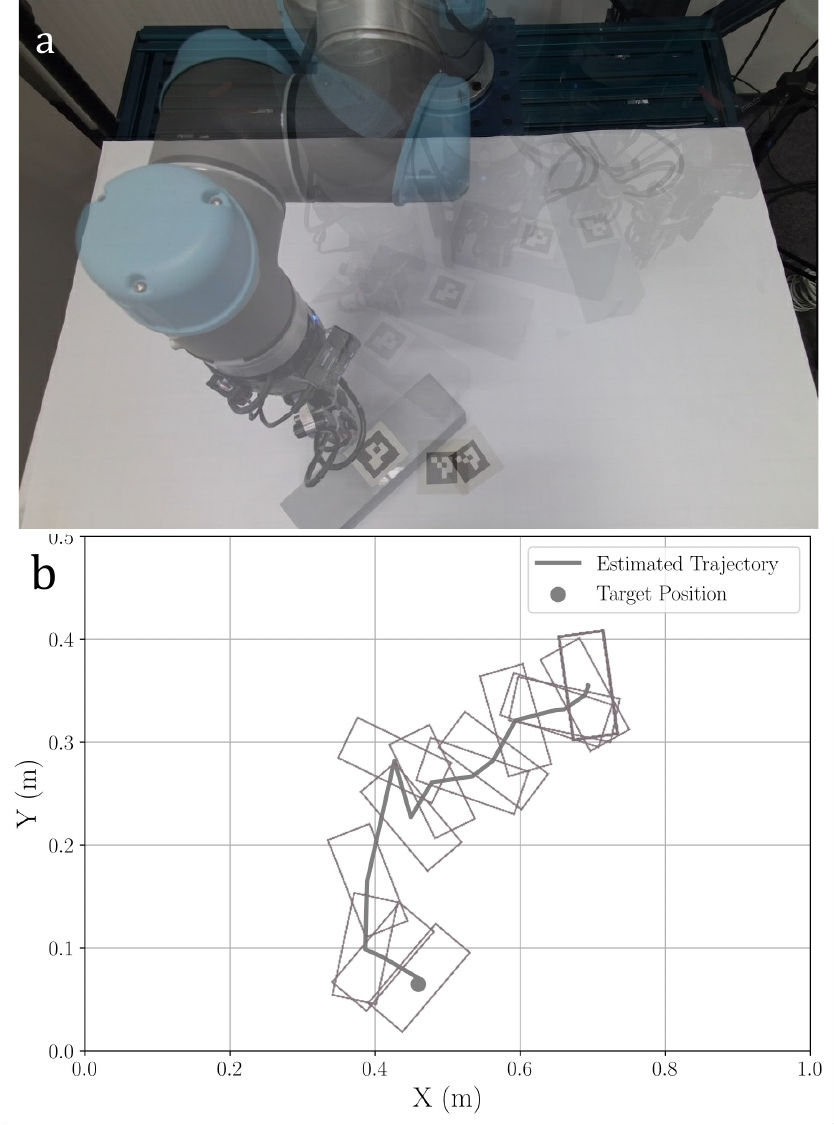}
    \caption{Goal-driven closed loop control of single object}
    \label{fig:goal_driven_single_obj}
\end{figure}


\begin{table*}[]
\centering
\caption{Goal-driven Closed Loop Control}
\label{tab:closed-loop-control-results}
\begin{tabular}{@{}llc@{}}

\toprule
\textbf{\begin{tabular}[c]{@{}l@{}}Experimental\\ Condition\end{tabular}} & \textbf{Object Type} & \textbf{Euclidean Error (cm)} \\ \midrule \midrule
\multirow{3}{*}{\textbf{Standard}}                                                                    & Articulated Revolute Object  &  2.302 $\pm$ 0.71 \\
\textbf{}                                                                  & Articulated Prismatic Object &  1.031 $\pm$ 0.96 \\
\textbf{}                                                                  & Single Object                &  1.908 $\pm$ 0.86\\ \midrule
\multirow{2}{*}{\textbf{\begin{tabular}[c]{@{}l@{}}Varied Center-\\ of-mass\end{tabular}}} & Articulated Object           &  3.570 $\pm$ 1.04\\
\textbf{}                                                                  & Single Object                & 3.604 $\pm$ 0.43  \\ \midrule
\multirow{2}{*}{\textbf{\begin{tabular}[c]{@{}l@{}}Challenging \\ Background\end{tabular}}} & Articulated Object           &  3.816 $\pm$ 0.97\\
\textbf{}                                                                  & Single Object                & 3.101 $\pm$ 0.24 \\ \midrule
\multirow{2}{*}{\textbf{Low light}}                                                        & Articulated Object           &  3.677 $\pm$ 0.43\\
\textbf{}                                                                  & Single Object                & 2.974 $\pm$ 0.92 \\ \bottomrule 
\end{tabular}
\end{table*}

\subsection{Closed-loop control for Goal-driven Manipulation}
\label{ssec:goalManip}
Since the closed-loop control requires accurate pose tracking, we perform extensive evaluations on various configurations of articulated objects with revolute and prismatic joints as well as for single objects. The target pose is chosen by a human user by placing the object in desired configuration at any arbitrary location in the robot workspace. A single RGB-D image is captured at the goal position and the part segmentation is performed to record the number of parts to track and the pose is recorded as the target pose. The object is then moved into an arbitrary initial pose in the robot workspace by the human user and the robot is tasked to manipulate the object into the goal-pose configuration. For single as well as articulated objects with revolute joints, the robot relies upon non-prehensile pushing manipulation for moving the object into a desired goal state and for articulated objects with prismatic joints, the hold-pull maneuver is used instead. The accuracy of the goal-driven manipulation for an object containing $N$ parts is computed by the $L^2$ norm in the X-Y plane with the final manipulated configuration of each part of the object with the corresponding goal pose as:

\begin{equation}
    ||e||_2 = \sqrt{(x - x_g)^2_i+(y - y_g)^2_i} \quad \text{for} \ i=1,2,\dots,N.
    \label{eq:l2norm}
\end{equation}
\begin{figure}[t!]
    \centering
    \includegraphics[width = 0.7\textwidth]{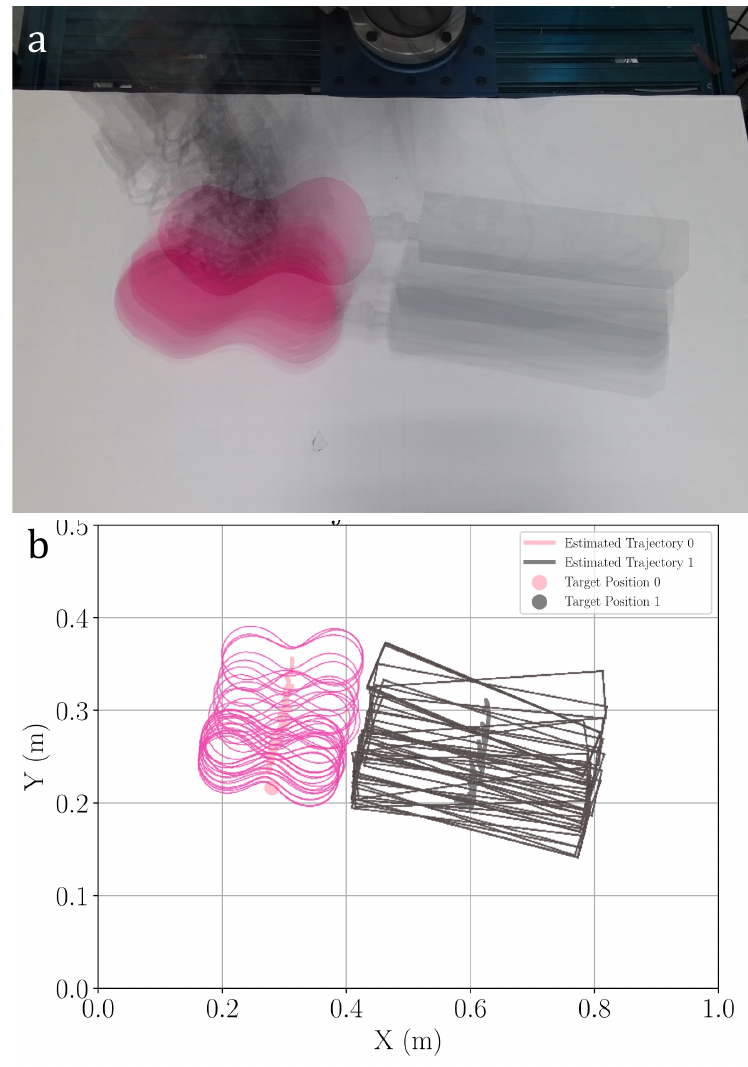}
    \caption{Goal-driven closed loop control of articulated object}
    \label{fig:goal_driven_articulated_obj}
\end{figure}
Consequently, for a single rigid object $N=1$ in the Eq.~\eqref{eq:l2norm}.
For single and articulated (revolute) objects, the robot performs iterative incremental pushing of each part one-by-one such that the distance to goal is minimized. This can be visualized in Fig.~\ref{fig:goal_driven_single_obj} for single objects and Fig.~\ref{fig:goal_driven_articulated_obj} for articulated objects with revolute joints. Numerical results are shown in Tab.~\ref{tab:closed-loop-control-results}. We note that the average error at goal-state for all objects (single/ articulated) is $<3$ cm. We performed five repeated trials for the four single objects and all revolute articulated objects (except \texttt{ blue-sine / green butter}) shown in Fig.~\ref{fig:object-list}, in total 55 repeated trials for this experiment. 

\begin{figure}[t!]
    \centering
    \includegraphics[width = \textwidth]{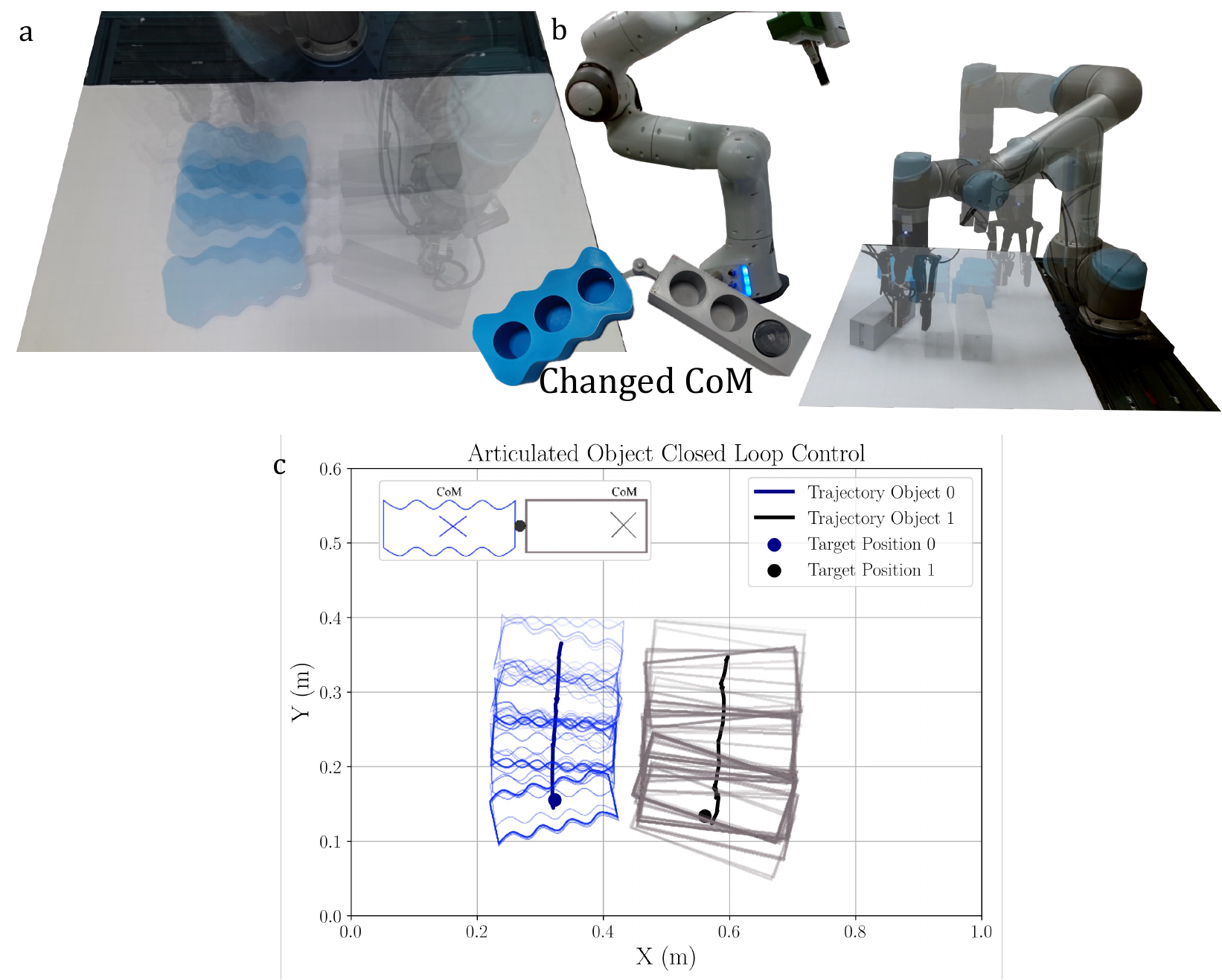}
    \caption{Goal-driven Pushing with varied Center of Mass: Articulated Object}
    \label{fig:com_articulated}
\end{figure}
\begin{figure}[t!]
    \centering
    \includegraphics[width = \textwidth]{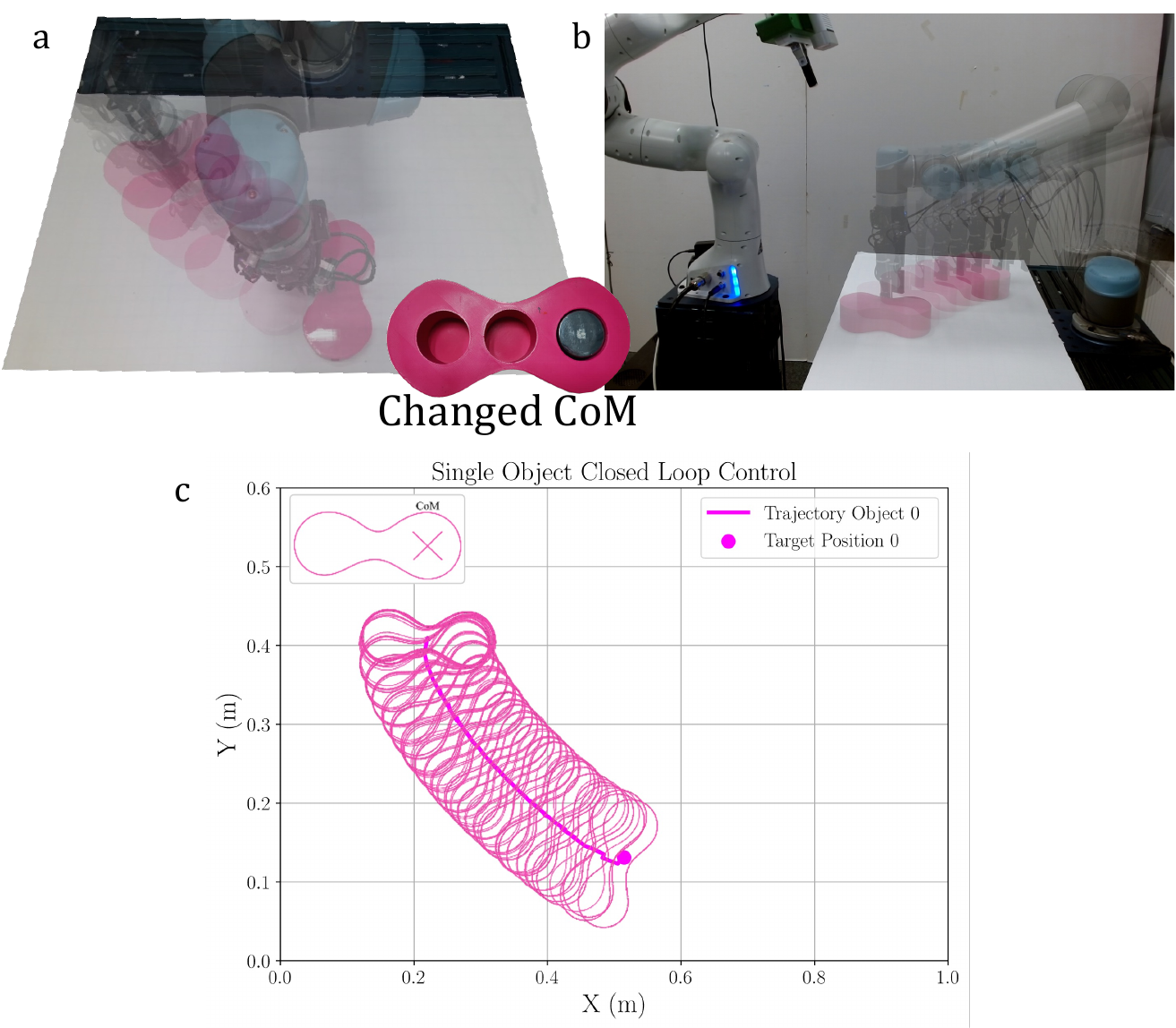}
    \caption{Goal-driven Pushing with varied Center of Mass: Single Object}
    \label{fig:com_single}
\end{figure}

\subsubsection{Goal-driven Manipulation for Prismatic Articulated Objects}
For articulated objects with prismatic joints, the desired pose is specified by the user by translating one segment of the articulated object to an arbitrary distance along the axis of articulation. Once the goal-pose of each part is extracted by using the part segmentation method, the user moves the parts back to an arbitrary initial pose along the axis of articulation. The task of the robot is to identify and track the current pose and manipulate the object into the goal pose. The hold-pull manipulation maneuver is used as it is ideal for manipulating such prismatic articulated objects. The Panda robot which is equipped with the RGB-D camera on its end-effector (in an \textit{eye-in-hand} configuration) performs the hold maneuver and the UR5 robot performs the grasp and pull maneuver. While the Panda robot is performing the holding maneuver, the object is very close to the camera and object tracking stops. Hence, for the grasp and pull maneuver, the UR5 robot relies upon the grasp pose detected prior to interaction. However, the tactile sensors allows the UR5 robot to adapt to minor grasp pose error. The gripper opens to the maximum limit and closes gradually until the tactile sensors detects contact. The UR5 robot performs a pull maneuver along the axis of articulation to the goal-pose. The pulling distance is calculated as the difference between goal pose and current pose. During the pull maneuver, the 3-axis contact force is monitored from each taxel in the tactile sensor array to detect possible loss of contact. Furthermore, a constant force of 5N is exerted by the gripper on the object. We note that the grasping force is sufficient to overcome friction in the prismatic joint. 
Ten repeated trials were executed for each of the two prismatic objects depicted in Fig.~\ref{fig:object-list}, totalling 20 repeated trials.
The average error of the prismatic articulated objects ($1.03 \pm 0.96$ cm) is less than that of revolute articulated objects ($2.30 \pm 0.71$ cm) due to limited displacement from the initial configuration possible with the prismatic joint.

\subsubsection{Goal-driven Pushing with Varied Center of Mass}

Typically we assume that the center of mass of an object coincides with the geometric centroid which can be computed as the mean of all the 3D points representing the object. However, in many scenarios this assumption fails to hold true. Hence, we design an experiment wherein an additional weight of 500g in the form of a calibrated weight metal cylinder is inserted into the object. The objects are 3D printed with holes embedded within which are capable of housing external objects. Depending on the object shape, the weight of individual objects vary between 200g-400g. Hence, the additional weight shifts the center of mass (CoM) of the object from the geometric centroid. However, through visual inspection alone the object with varied center of mass is indistinguishable from an identical shaped object without the additional weight inserted into it. This experiment evaluates the ability of our approach to push the object(s) to the goal configuration with varied CoM. The shifted CoM causes the object to turn when pushed through the geometric center. While inferring the CoM position through interaction goes beyond the scope of this work, the robot relies upon visual and tactile feedback to compensate for the undesired motions caused during pushing. This is shown in Fig.~\ref{fig:com_articulated} for pushing articulated revolute object wherein the CoM of the grey \texttt{cuboid} object is shifted with 500g placed on the right hole. As seen from Fig.~\ref{fig:com_articulated}c, the robot is capable of pushing the articulated object to the goal state. 
We performed five repeated experiments for the external weight placed in each of the three holes for the \texttt{pink-butter} single object and three repeated trials for each of the six holes for \texttt{blue-sine/gray-cuboid} articulated object resulting in a total of 33 repeated trials.
We obtain an average error of $3.816 \pm 0.97 $ cm from the repeated trials. A similar result is also seen with a singular object with shifted center of mass while pushing to goal pose (exemplary case shown in Fig.~\ref{fig:com_single}) with the average error at the goal-state being $3.604 \pm 0.43$ cm.

\begin{figure}[t!]
    \centering
    \includegraphics[width = 0.7\columnwidth]{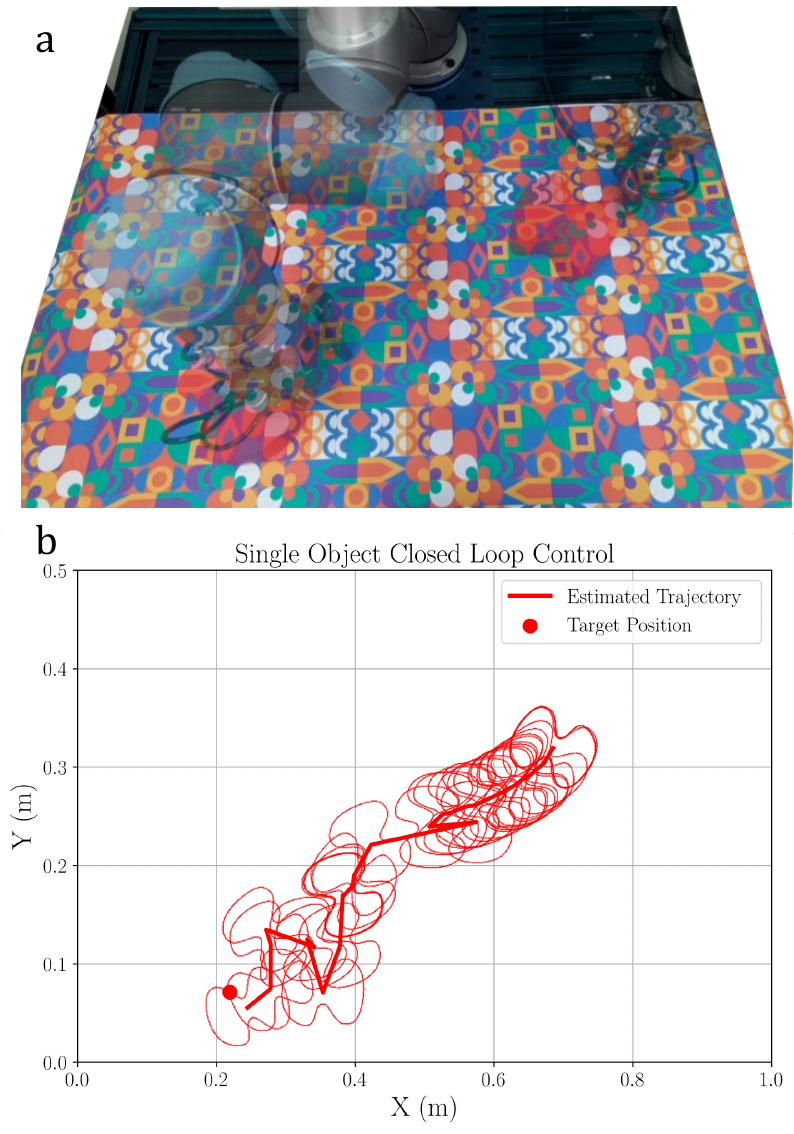}
    \caption{Goal-driven Pushing with challenging background: Single Object}
    \label{fig:color_back_single}
\end{figure}
\begin{figure}[t!]
    \centering
    \includegraphics[width = 0.7\columnwidth]{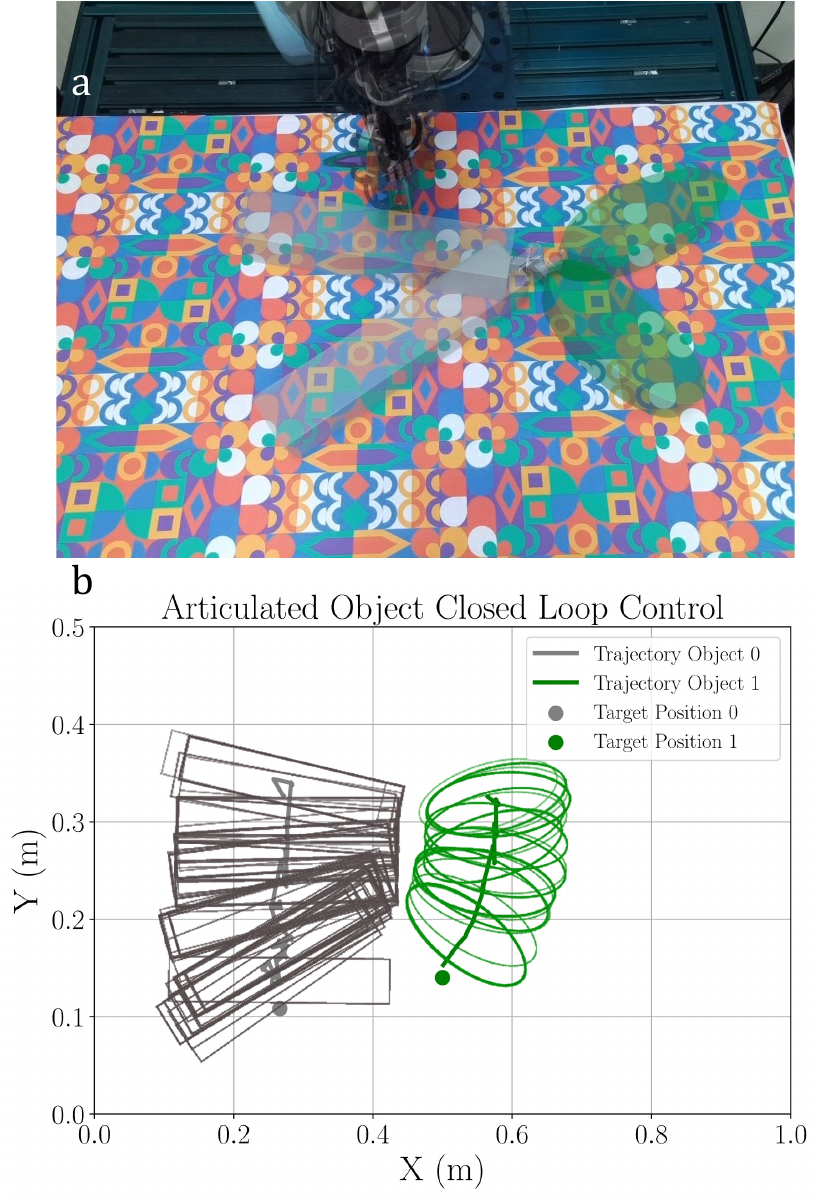}
    \caption{Goal-driven Pushing with challenging background: Articulated Object}
    \label{fig:color_back_single_articulated}
\end{figure}
\subsubsection{Goal-driven Pushing with Challenging Background}
As our part-segmentation approach relies upon visual point clouds and separating the foreground (object) from background points, we perform an experiment with a challenging colored pattern background with the articulated and single objects for goal-driven pushing as seen in Fig.~\ref{fig:color_back_single} and Fig.~\ref{fig:color_back_single_articulated}.
To increase the complexity, we also intentionally used a revolute articulated object with all parts having the same colour. Although our part segmentation approach is based on RGB point cloud data for the region growing method, we demonstrate that it is unaffected by challenging coloured backgrounds as seen by the object trajectory and proximity to the target pose in Fig.~\ref{fig:color_back_single}b, Fig~\ref{fig:color_back_single_articulated}b.
We performed a total of 20 repeated trials for single and articulated objects respectively.
The average error for target-driven control is $3.816 \pm 0.97$ cm for articulated objects whereas it is $3.1 \pm 0.24$ cm for single objects with challenging backgrounds. The discrepancy in error is approximately $1.5$ cm greater, on average, in comparison to the corresponding objects presented against a standard white-colored background. We note that the challenging background color does not adversely affect our ArtReg tracking algorithm and closed-loop control approach.

\begin{figure}[t!]
    \centering
    \includegraphics[width = 0.7\textwidth]{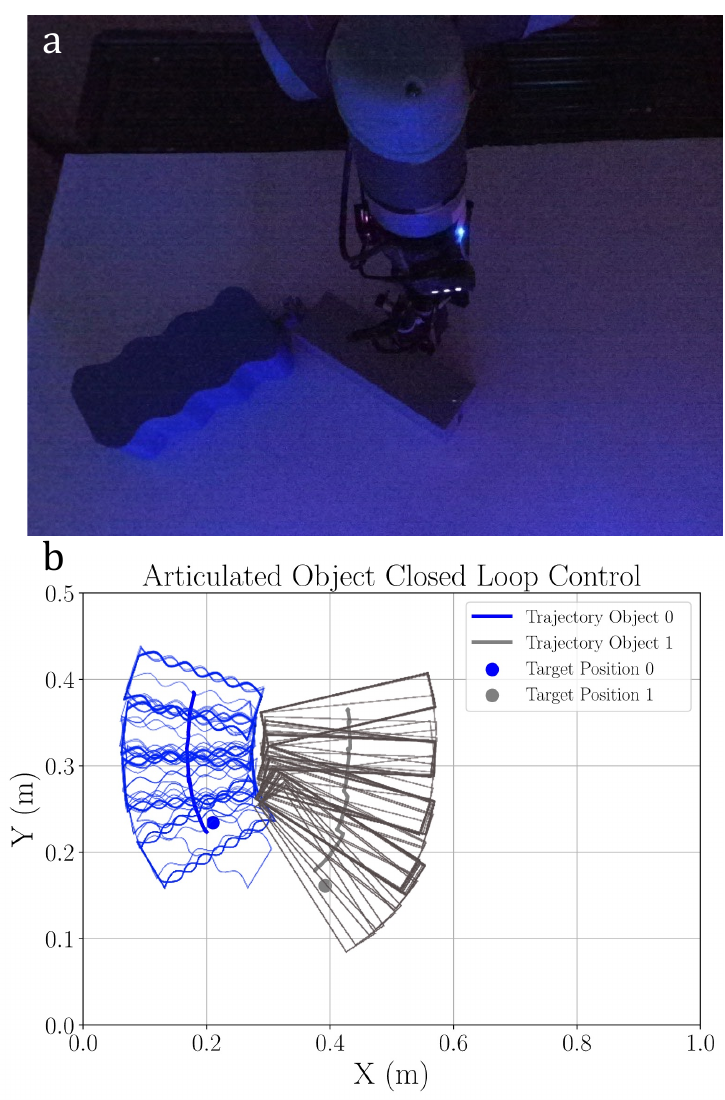}
    \caption{Goal-driven Pushing with low ambient light conditions: Articulated Object}
    \label{fig:low_light_articulated_articulated}
\end{figure}
\subsubsection{Goal-driven Pushing in Low-Light Conditions}
Furthermore for robustness testing, we performed goal-driven pushing experiments with single objects and articulated objects in low light conditions. It is well known that vision-based methods are susceptible to diminished lighting conditions. During the experiments, we deactivated the overhead lights in the room and minimal light was emanating from the computer screen that was used to control the robot as seen in Fig.~\ref{fig:low_light_articulated_articulated}a. While the part segmentation method relies upon visual point clouds, the robot interaction relies upon tactile sensing as well. In instances where an erroneous pushing configuration is identified, the robot upon detecting unintended contact during movement towards the designated pose, promptly halts its operation and reverts to its initial position. By leveraging vision and tactile sensing the robot is able to push the objects to goal within the prescribed margin of error as seen the trajectory shown in Fig.~\ref{fig:low_light_articulated_articulated}b. 
We performed a total of 20 repeated trials for single and articulated objects respectively.
The average error in low light conditions for articulated objects is $3.67 \pm 0.43$ cm and for single objects is $2.97 \pm 0.92$ cm.

\subsection{Object Tracking}
\label{ssec:tracking_exp}
To isolate and precisely assess the accuracy of the ArtReg tracker without any error propagation stemming from the closed-loop controller, the robot is manually operated via the teach pendant to maneuver the object through a series of arbitrary poses. Furthermore, we also performed additional experiments wherein multiple objects were moved in the workspace by the user in randomised trajectories.
Aruco markers are placed on each object which are used to gather the ground-truth poses. The average euclidean distance error calculated at 5 frames-per-second from Eq.~\eqref{eq:l2norm} is presented in Tab.~\ref{tab:obj-tracking-results}. 
We performed 10 repeated trails for each type of tracking experiment: single, multiple and articulated. 
For all objects (single, multiple and articulated), the average pose tracking error with our ArtReg algorithm is less than $2$ cm. The least tracking error is achieved for single objects as expected with an average error of $1.313 \pm 0.27$ cm whereas the tracking error increases marginally for articulated objects (where 2 objects are tracked) with an average error of $1.406 \pm 0.48$ cm and for multiple objects (with the object number $N$ ranging from $3 \leq N \leq 7$) with an average tracking error of~$1.512 \pm 0.31$ cm.

\begin{table}[]
\centering
\caption{Object Tracking}
\label{tab:obj-tracking-results}
\begin{tabular}{@{}lc@{}}
\toprule
\textbf{Object Type} & \textbf{Euclidean Error (cm)} \\ \midrule \midrule
 Single Objects  &   1.313 $\pm$ 0.27\\
 Multiple Objects &  1.512 $\pm$ 0.31 \\
 Articulated Objects & 1.406 $\pm$ 0.48  \\ 
\bottomrule 
\end{tabular}
\end{table}

\subsection{Baseline Comparison}
\label{ssec:comps_exp}
\begin{figure}[t!]
    \centering
    \includegraphics[width = 0.6\textwidth]{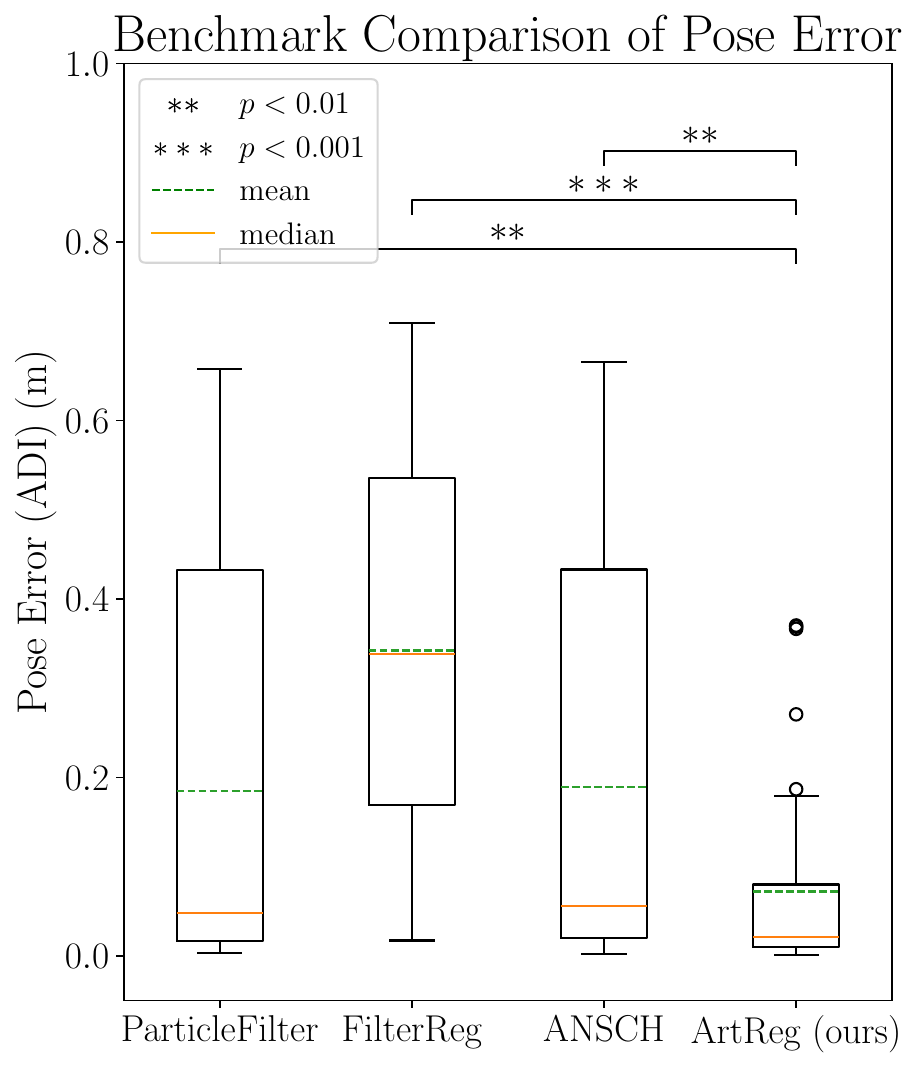}
    \caption{Pose estimation results with simulated articulated objects from PartNet-mobility dataset in random pose configurations with comparisons against state-of-the-art. $p$ values calculated by Welch's t-test shown as $\ast$.}
    \label{fig:benchmark_results}
\end{figure}

\begin{figure}[t!]
    \centering
    \includegraphics[width = 0.6\textwidth]{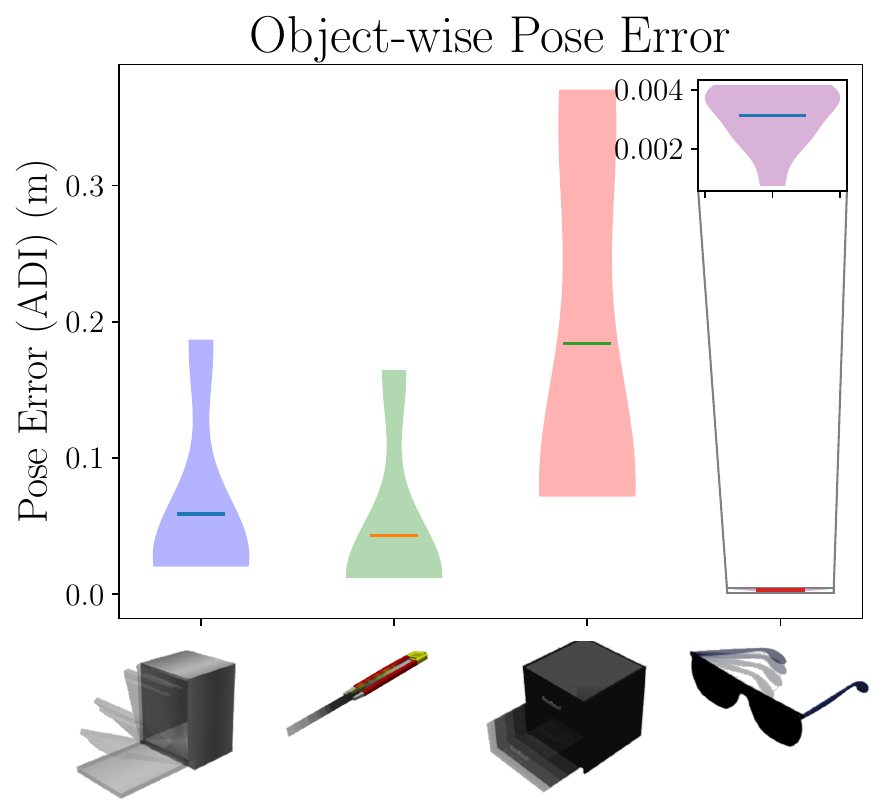}
    \caption{Violin-plots showing the object-wise pose estimation results with our ArtReg method. The notches in the violin-plot shows the mean value.}
    \label{fig:object_wise_results}
\end{figure}

In this section, we compare our ArtReg tracking algorithm with the following state-of-the-art methods: (a) particle filter-based tracker that has been used in many prior works~\citep{cabido2009high, gonzales2015combinatorial, cifuentes2016probabilistic}; (b) learning-based approach termed ANSCH from Li et al.~\citep{li2020category} and (c) FilterReg algorithm presented in~\citep{gao2019filterreg}.
We use a standard benchmark dataset for comparison termed as the PartNet-Mobility dataset~\citep{Xiang_2020_SAPIEN}. The PartNet-Mobility dataset consists of various real-world articulated mesh models with movable part definitions. We choose the models from the following categories: dishwasher (1 DoF revolute joint), glasses (two 1 DoF revolute joints), drawer (1 DoF prismatic joint), and blade (1 DoF prismatic joint). A sample of the chosen category of articulated objects are shown in Tab.~\ref{tab:benchmark_sim}. The object models are provided as URDF files which are simulated in a PyBullet simulator~\citep{coumans2019}.
For all the objects, the joint values are in the range $[0,1]$.
We randomly sample 10 different poses for each object by choosing different values for the joint angle/ distance in the range $[0,1]$ and articulating the moving link according to the joint value. For each pose, a point cloud is captured by a simulated depth camera at a 45$\degree$ top-down viewing angle. This results in partial point clouds for the objects as seen from Tab.~\ref{tab:benchmark_sim} and provides a challenging benchmark for all the baseline methods as well as our approach.
As we focus on the tracking accuracy, we provide the model point clouds of each link as input that are connected by the articulation joints that is directly sampled from the CAD mesh files. 
\textbf{Baseline Implementation Details:} We used the implementation of the FilterReg algorithm~\citep{gao2019filterreg} available in the Python package \texttt{probreg}~\citep{probreg}, ANSCH method from official GitHub implementation~\citep{li2020category} and particle filter re-implemented using Open3D modules~\citep{Zhou2018} for point cloud processing. All baseline approaches are implemented in Python.
For quantitative evaluation, we use the Average Distance of
model points with Indistinguishable views metric (ADI) which is insensitive to object symmetries~\citep{hinterstoisser2012model}. The ADI metric is measured as follows:
\begin{equation}
    \mathtt{err}_{adi} = \frac{1}{|\mathcal{O}|}\sum_{\mathbf{p}_1 \in \mathcal{O}} \min_{\mathbf{p}_2 \in \mathcal{O}} || (\mathbf{R}_{gt}\mathbf{p}_1 + \mathbf{t}_{gt}) - (\mathbf{R}_{est}\mathbf{p}_2 + \mathbf{t}_{est}) ||,
    \label{eq:adi}
\end{equation}
where $(\mathbf{R}_{gt}, \mathbf{t}_{gt})$ and $(\mathbf{R}_{est}, \mathbf{t}_{est})$ refers to ground-truth and estimated rotation and translation respectively, $\mathcal{O}$ refers to the object model point cloud and the points $p_1 \in \mathcal{O}$ and $p_2 \in \mathcal{O}$ belong to the object point cloud and denote the closest corresponding points when $\mathcal{O}$ is transformed by $\{\mathbf{R}_{gt}, \mathbf{t}_{gt}$\} and $\{\mathbf{R}_{est}, \mathbf{t}_{est}$\} respectively. The quantitative comparison against baseline approaches is presented in Fig.~\ref{fig:benchmark_results}. Furthermore, object-wise results are presented in Fig.~\ref{fig:object_wise_results}. 
Qualitative pose estimation results for selected poses are presented in Tab.~\ref{tab:benchmark_sim} and discussed in the following section.

\begin{table*}[]
\centering
\caption{Examples of simulated articulated objects from PartNet-Mobility dataset~\citep{Xiang_2020_SAPIEN} used for benchmarking our ArtReg method against state-of-the-art algorithms.}
\label{tab:benchmark_sim}
\resizebox{\textwidth}{!}{%
\begin{tabular}{@{}lcccc@{}}
\toprule
                                                                     &\textbf{Dishwasher}                                            & \textbf{Glasses}                                               & \textbf{Drawer}                                                 & \textbf{Blade}                                                  \\ \midrule
\textbf{Object}                                                      &       \parbox[c]{3.5cm}{
\includegraphics[height=3cm]{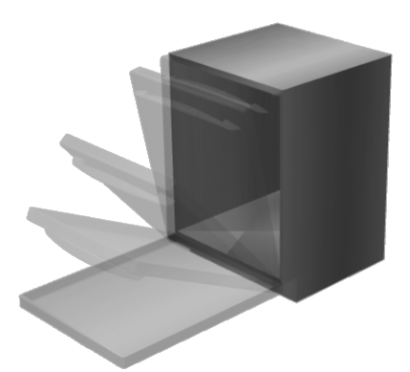}}                                                        &     \parbox[c]{3.5cm}{
\includegraphics[height=3cm]{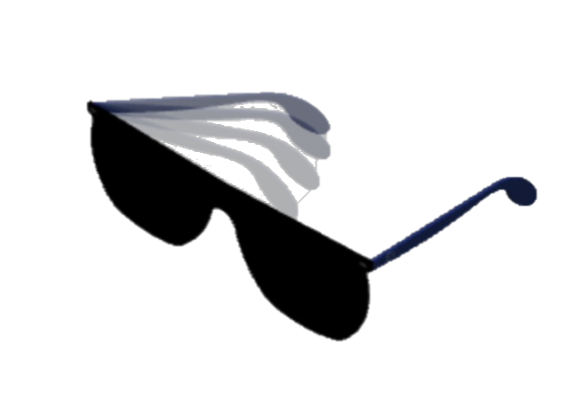}}                                                            &            \parbox[c]{3.5cm}{
\includegraphics[height=3cm]{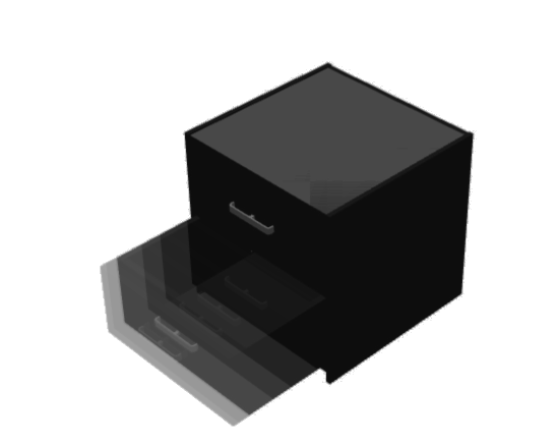}}                                                      &                                   \parbox[c]{3.5cm}{
\includegraphics[height=2cm]{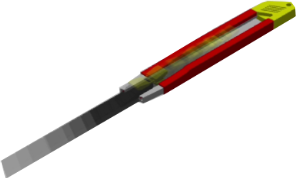}}                               \\ \\
\textbf{\begin{tabular}[c]{@{}c@{}}Articulation\\ Type\end{tabular}} & \begin{tabular}[c]{@{}l@{}}1 DoF Revolute\\ Joint\end{tabular} & \begin{tabular}[c]{@{}l@{}}2 DoF Revolute\\ Joint\end{tabular} & \begin{tabular}[c]{@{}l@{}}1 DoF Prismatic\\ Joint\end{tabular} & \begin{tabular}[c]{@{}l@{}}1 DoF Prismatic\\ Joint\end{tabular} \\
\textbf{Measured point cloud} & \parbox[c]{3.5cm}{
\includegraphics[height=3cm]{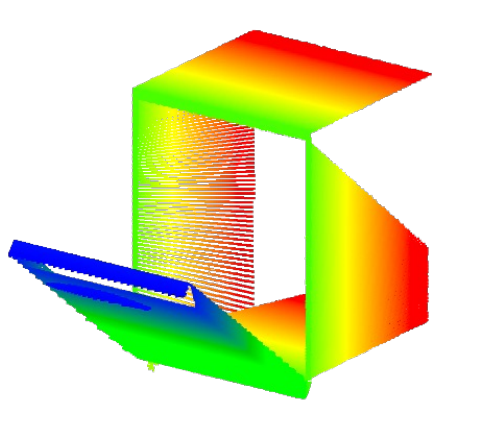}} & \parbox[c]{3.5cm}{
\includegraphics[height=3cm]{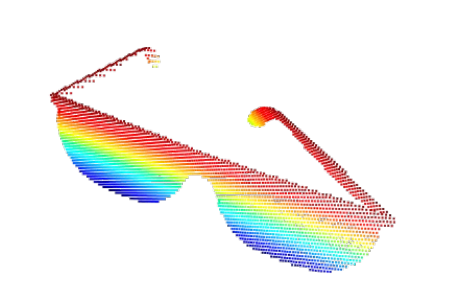}} &  \parbox[c]{3.5cm}{
\includegraphics[height=3cm]{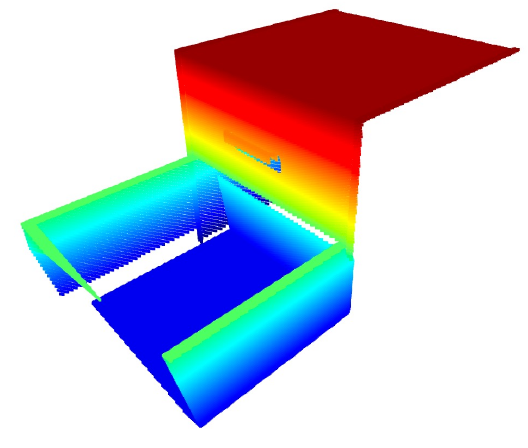}} & \parbox[c]{3.5cm}{
\includegraphics[height=1.8cm]{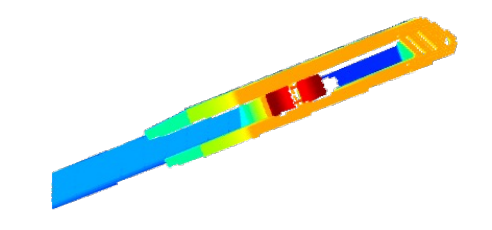}}\\
\textbf{Particle Filter} & \parbox[c]{3.5cm}{
\includegraphics[height=3cm]{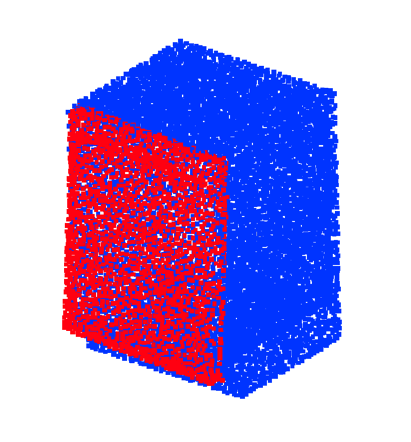}}& \parbox[c]{3.5cm}{
\includegraphics[height=3cm]{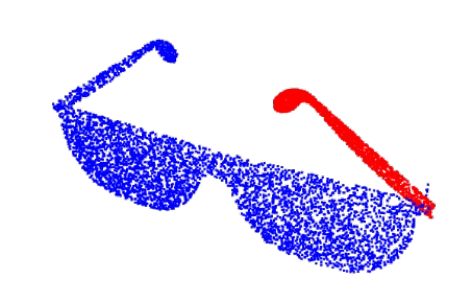}} & \parbox[c]{2.8cm}{
\includegraphics[height=3cm]{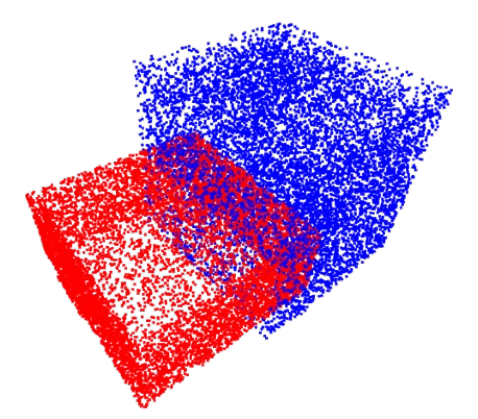}}  & \parbox[c]{3.5cm}{
\includegraphics[height=1.8cm]{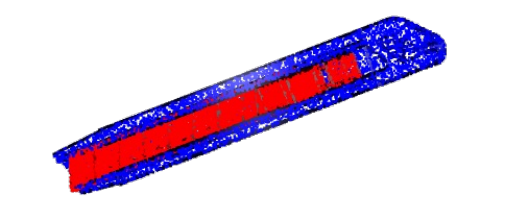}}\\
\textbf{FilterReg}~\citep{gao2019filterreg} & \parbox[c]{3.5cm}{
\includegraphics[height=3cm]{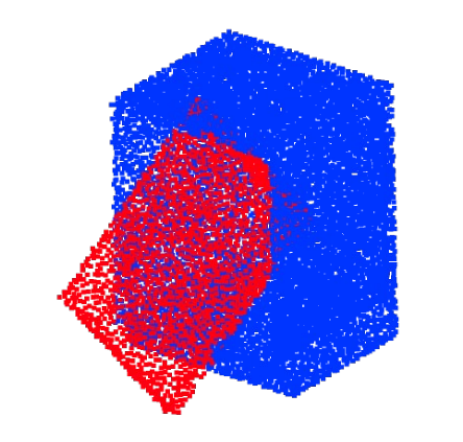}}&  \parbox[c]{3.5cm}{
\includegraphics[height=3cm]{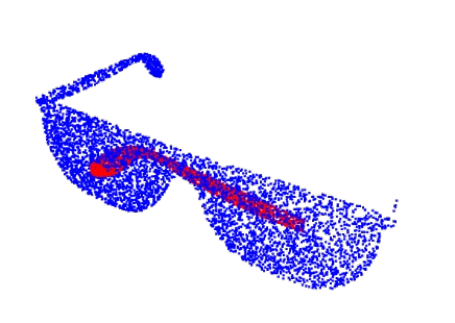}}&  \parbox[c]{3.5cm}{
\includegraphics[height=3cm]{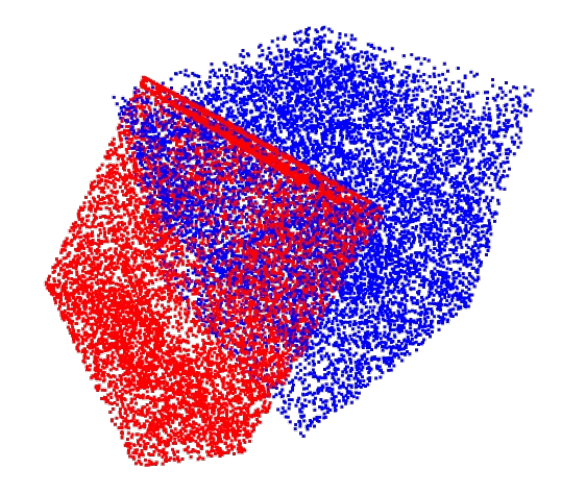}} & \parbox[c]{3.5cm}{
\includegraphics[height=1.5cm]{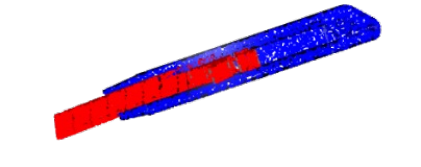}}\\
\textbf{ANSCH}~\citep{li2020category} & \parbox[c]{3.5cm}{
\includegraphics[height=3cm]{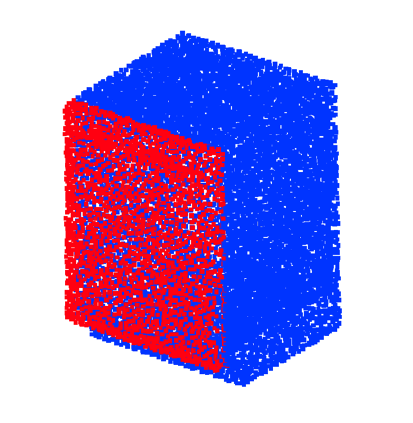}}& \parbox[c]{3.5cm}{
\includegraphics[height=3cm]{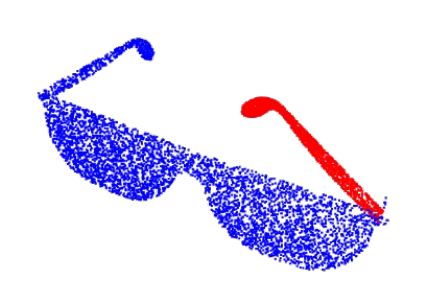}}&  \parbox[c]{3.5cm}{
\includegraphics[height=3cm]{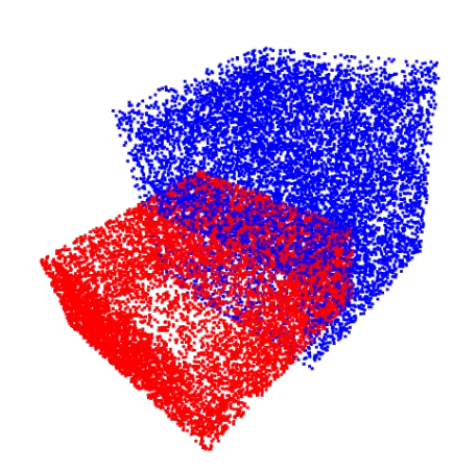}} & \parbox[c]{3.5cm}{
\includegraphics[height=1.8cm]{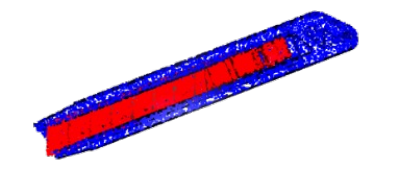}}\\
\textbf{ArtReg (ours)} &\parbox[c]{3.5cm}{
\includegraphics[height=3cm]{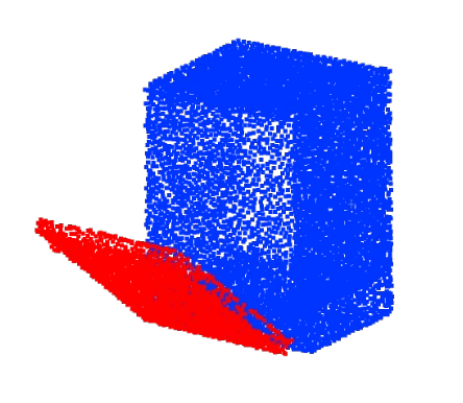}} & \parbox[c]{3.5cm}{
\includegraphics[height=3cm]{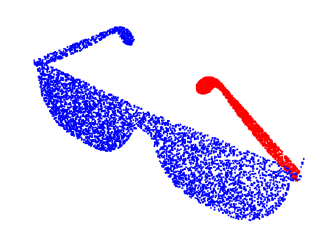}}&  \parbox[c]{3.5cm}{
\includegraphics[height=3cm]{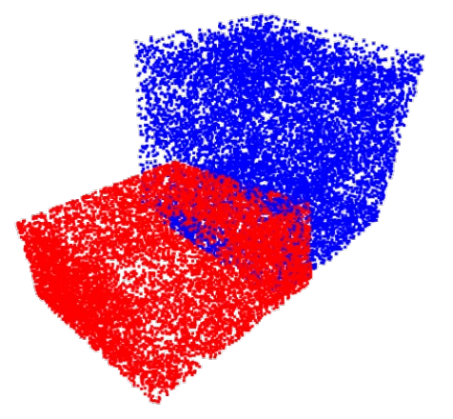}} & \parbox[c]{3.5cm}{
\includegraphics[height=1.8cm]{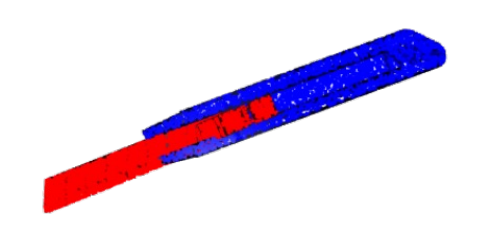}}
\\\bottomrule
\end{tabular}
} %
\end{table*}

\subsection{Discussion}
In this work, we presented a full-fledged framework for articulated object detection, pose estimation and tracking as well as goal-driven closed-loop control. Our ArtReg algorithm has been demonstrated to perform accurate and robust pose tracking of single, multiple, and articulated objects. As illustrated in Fig.~\ref{fig:benchmark_results}, it is evident that the ArtReg algorithm outperforms the state-of-the-art methodologies: our approach demonstrates an average accuracy improvement of approximately $60\%$ and a reduction in median error exceeding $50\%$ compared to baseline methods ($p<0.01$ as determined by Welsh's t-test). The mean ADI error value for our ArtReg algorithm is $7.23$ cm and median ADI error is $2.1$ cm for all the simulated objects. The object-wise pose results in Fig.~\ref{fig:object_wise_results} demonstrates that our method provides highly accurate estimates for articulated objects with completely measured point clouds without self-occlusions (such as glasses) with average ADI error of $0.312$ cm and relatively larger errors for articulated objects such as drawer (average ADI error $18.4$ cm) due to high self-occlusion as well as the larger size (bounding box size upto 2 m). The superiority of our methodology over state-of-the-art techniques is qualitatively evident in Tab.~\ref{tab:benchmark_sim}. Specifically, existing methods demonstrate a deficiency in accurately capturing the pose of articulated objects, such as dishwashers or drawers, whereas our approach yields precise estimations. Furthermore, our ArtReg approach does not require any prior knowledge of the object whereas the recent state-of-the-art approaches such as ANSCH~\citep{li2020category} require prior training on a large labelled dataset of category-level object with articulation and pose information.    

Considering real-world robotic experiments, our framework achieves highly accurate pose tracking ($<1.5$ cm average error over the trajectory) and goal-driven closed-loop control ($<4$ cm average error at goal-state). We demonstrated the robustness of our approach in various conditions such as low ambient light, challenging backgrounds, and varying the center-of-mass of the objects. We note the importance of tactile perception in our framework: for the detection of articulated objects, where objects have identical visual features (such as Fig.~\ref{fig:prism_action_detection}), robots rely on tactile feedback during interactive perception to discern the articulated object. During goal-driven manipulation, the UR5 robot relies on tactile force feedback to ensure contact during manipulation, and in case of loss of contact, the framework re-triggers the computation of the push or hold-pull pose to perform the manipulation task until the goal-state is achieved. Furthermore, tactile feedback is crucial in situations such as low-light conditions, and when the center-of-mass (CoM) of the object is varied by the user. Although we do not directly infer the CoM location, the tactile feedback during the manipulation maneuvers and the resulting visual feedback of the effect of the manipulation allows the robot to adapt the pushing strategy such that the goal-pose is achieved. The synergistic combination of visual and tactile perception allows the two robots equipped with the complementary sensing abilities to detect, track and manipulate various types of objects in a robust manner.

\section{Conclusions}
\label{sec:conclusion}
In this work, we presented a novel SE(3) Lie Group-based Unscented Kalman Filter approach for real-time object tracking termed ArtReg. 
Our approach is demonstrated to perform robustly and accurately with various types of objects such as single, multiple, and articulated objects under different conditions such as low-light, challenging backgrounds and varying center-of-mass. Visual and tactile perception are seamlessly integrated in our full-fledged framework which allows the robots to detect possible kinematic articulation in objects using interactive manipulation, track the object using our ArtReg tracker and perform closed-loop goal-driven manipulation to bring the objects to the desired goal-state. 
The two-robot team equipped with visual and tactile sensing, respectively, performed various types of manipulation maneuvers such as pushing or hold-pulling depending on the type of articulated object (revolute or prismatic joints) to perform goal-driven manipulation.
Our ArtReg algorithm is also benchmarked against various state-of-the-art approaches, and we demonstrated outperformance on a standard benchmark dataset consisting of various types of articulated objects. 

Since our ArtReg algorithm works on point cloud data, it can readily be combined with learning-based approaches providing semantic segmentation of measured point cloud data to improve tracking in complex scenarios. While we focused on popular joint types such as revolute and prismatic joints, our framework for tracking and manipulation of articulated objects can be extended for articulated objects with other types of joints such as universal joint, screw joints, ball joint, and so on which are considered as future work.

\bibliographystyle{elsarticle-num} 
\bibliography{root}

\end{document}